\documentclass[runningheads]{llncs}
\usepackage{graphicx}
 
\usepackage{eccv}

\usepackage{lmodern}
\usepackage[T1]{fontenc}

\usepackage{eccvabbrv}

\usepackage{amsmath}
\usepackage{booktabs}
\usepackage{enumitem}
\usepackage{multirow}
\usepackage[accsupp]{axessibility}  
\usepackage[export]{adjustbox}

\usepackage{xcolor} 

\usepackage[
  colorlinks=true,
  linkcolor=eccvblue,
  citecolor=eccvblue,
  urlcolor=eccvblue,
  pdfborder={0 0 0}
]{hyperref}

\usepackage{etoc}

\usepackage{orcidlink}

\begin{document}


\newcommand{\titleicon}{%
  \raisebox{-0.2cm}{\includegraphics[height=0.7cm]{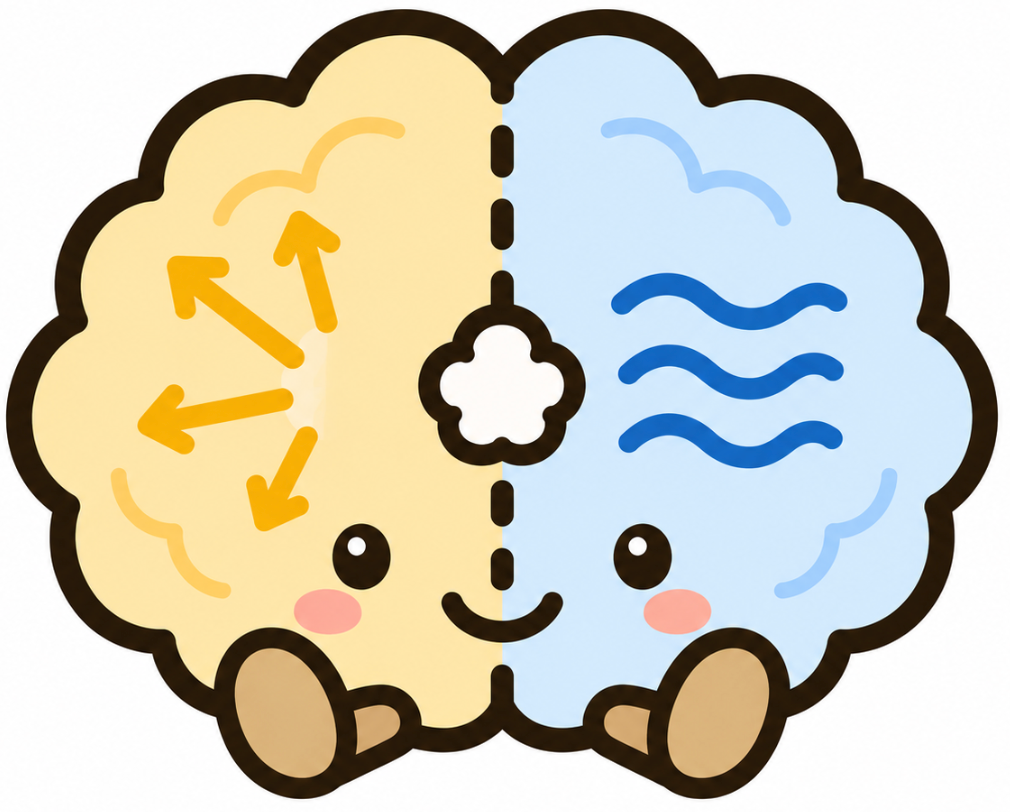}}%
}

\title{
    \hspace*{-1.3cm}
    \titleicon
    \hspace{0.06cm}
    Physics-Grounded Disentangled Flow Modeling for Brain Disease Progression Trajectory
}

\titlerunning{\textbf{PDF}: \textbf{P}hysics-Grounded \textbf{D}isentangled \textbf{F}low}

\author{Jun Wang \and Peirong Liu\thanks{Corresponding author.}}
\authorrunning{J. Wang and P. Liu}
%
\institute{Department of Electrical and Computer Engineering,\\
Data Science and AI Institute,\\
Johns Hopkins University\\ \vspace{0.15cm}
\email{\{jwang674,pliu53\}@jh.edu}
}

\maketitle

\sloppy

\begin{abstract}
Forecasting longitudinal brain lesion evolution is critical for disease monitoring and treatment planning. Existing approaches typically learn a direct mapping from a baseline image to a future observation, without explicitly modeling the physical mechanisms underlying the lesion progression. Such an entangled modeling of structural deformation and image intensity variation limits physical plausibility, model generalization, and interpretability. To address this, we propose \textbf{PDF}, a \textbf{P}hysics-grounded \textbf{D}isentangled \textbf{F}low matching framework for longitudinal brain disease forecasting. We explicitly decompose the longitudinal modeling of lesion growth into two processes, each learned by a dedicated flow matching network: morphology evolution, which captures lesion growth and structural deformation; and intensity evolution, which models signal changes driven by variations in lesion concentration. To enforce physics-grounded constraints, we introduce a PDE-regularized loss based on lesion growth dynamics, that enforces a diffusion-reaction-advection formulation for morphological evolution. Experiments on three public longitudinal datasets spanning diverse brain diseases demonstrate state-of-the-art performance, validating the effectiveness of the disentangled modeling framework and physics-grounded learning design. Code is publicly available at \url{https://github.com/jhuldr/PDF}.

\keywords{Physics-Informed Learning; Longitudinal Disease Modeling}
\end{abstract}

\section{Introduction}
\label{sec:intro}

\begin{figure}[t]
\centering
\includegraphics[width=1\linewidth]{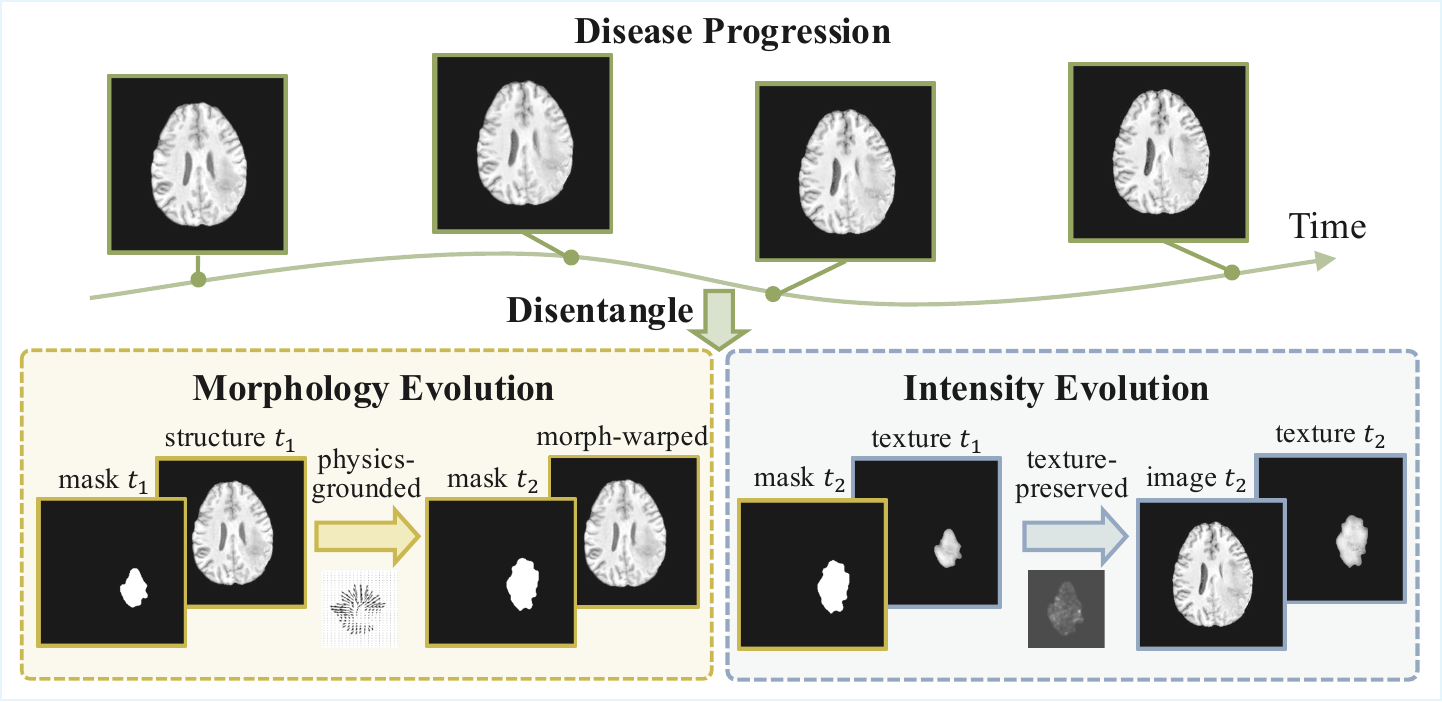}
\caption{\textbf{Disentangled modeling of brain disease progression.} 
\textbf{Top:} irregularly sampled longitudinal MRI observations illustrating disease progression trajectory over time. 
\textbf{Bottom:} overview of the proposed disentangled framework, where morphology evolution models lesion growth and associated structural deformation, while intensity evolution captures signal changes induced by lesion concentration variations.}
\label{fig:disentagle}
\end{figure}

Accurate prediction of longitudinal brain disease progression is clinically important, as anticipating lesion evolution supports treatment planning, therapy adjustment, and risk assessment~\cite{durso2022personalized, wang2020deep, stupp2014high, jackson2015patient}. 
Advances in imaging technologies have enabled routine longitudinal magnetic resonance imaging (MRI) acquisition, making it possible to model disease trajectories and forecast future structural changes~\cite{nabrawi2023imaging, liu2025imageflownet, lachinov2023learning}. 
Despite these advances, longitudinal prediction of disease progression remains challenging due to limited data, irregular time intervals between scans, and imaging or protocol variability across time points, which together hinder reliable and generalizable progression modeling~\cite{fonteijn2011event, ruan2025comprehensive, nguyen2023clinically}.

Recent advances in continuous-time and generative modeling have substantially improved longitudinal disease progression modeling, particularly through diffusion- and flow-based approaches~\cite{liu2023i2sb, liu2025imageflownet, zhang2024tfm, islam2026longitudinal, chen2026learning}.
I$^2$SB~\cite{liu2023i2sb} introduces a Schr\"odinger Bridge formulation within diffusion models to capture transitions between complex image distributions. 
ImageFlowNet~\cite{liu2025imageflownet} models disease evolution as a vector field learned through a position-parameterized neural ordinary differential equation (ODE), 
while trajectory flow matching~\cite{zhang2024tfm} trains neural stochastic differential equations in a simulation-free manner to model continuous trajectories.
Despite their flexibility, these methods often model progression as direct image-to-image translation from baseline. Such formulation implicitly entangles structural deformation and intensity variation without modeling the underlying physical mechanisms of lesion growth. As a result, they risk generating longitudinal predictions that lack physical consistency and reliability.

To address these limitations, we propose \textbf{PDF}, a \textbf{P}hysics-grounded \textbf{D}isentangled \textbf{F}low matching framework for longitudinal prediction of brain disease progression. 
Instead of directly generating a future scan, we decompose the brain disease progression into two complementary processes: morphological evolution and intensity evolution, as shown in Fig.~\ref{fig:disentagle}. 
The former captures lesion growth and its associated structural deformation, while the latter models signal changes driven by variations in lesion concentration. 
Each process is learned by a dedicated flow matching network, enabling structured and interpretable modeling of disease progression trajectories.

To enforce physically consistent lesion growth and improve the reliability of longitudinal prediction, we incorporate prior knowledge from lesion growth dynamics into morphological evolution modeling. 
Inspired by the lesion growth modeling through Fisher-KPP partial differential equation (PDE)~\cite{metzcar2024review, martens2022deep}, we propose a PDE-regularized loss that constrains the predicted velocity fields to satisfy the diffusion-reaction-advection formulation. We demonstrated that such physics-grounded constraints encourage biologically plausible structural changes and improve longitudinal consistency. Our main contributions are:
\vspace{-0.1cm}
\begin{itemize}
\item 
We propose a disentangled framework that explicitly decomposes brain disease progression into morphology evolution and intensity evolution, enabling interpretable and physically consistent modeling of disease dynamics.

\item 
We introduce a physics-grounded regularization based on the Fisher-KPP PDE, constraining the predicted flow fields to follow a diffusion–reaction–advection formulation, and thereby enforcing biologically plausible structural evolution.

\item  
We conduct experiments on three longitudinal brain MRI datasets with diverse modalities and diseases. Our method achieves state-of-the-art performance while producing more physically consistent predictions.

\end{itemize}

\section{Related Work}

\subsection{Generative Modeling for Longitudinal Disease Progression}
Generative and continuous-time models have recently become a dominant paradigm for modeling longitudinal aging analysis and disease progression~\cite{jian2026temporal,bai2024noder,marcus2010open}. 
Recent advances in diffusion and flow-based models enable flexible modeling of complex high-dimensional distributions over time~\cite{liu2023i2sb, puglisi2025brain, liu2025imageflownet, zhang2024tfm, islam2026longitudinal, chen2026learning, wang2025usb}.

Diffusion-based generative models~\cite{ho2020denoising,song2021scorebased} have demonstrated remarkable performance in image synthesis and have been widely adopted in medical imaging tasks. 
To better model transitions between distributions, I$^2$SB~\cite{liu2023i2sb} formulates image-to-image translation as learning a Schr\"odinger bridge between endpoint distributions, allowing more direct modeling of temporal transitions beyond standard Gaussian-to-data diffusion.
Such bridge-based perspectives provide a principled way to connect observations across time.

Flow-based continuous-time formulations offer a natural alternative for trajectory modeling.
Neural ODEs~\cite{chen2018neural} and neural SDEs~\cite{li2020scalable} parameterize temporal dynamics through differential equations, accommodating irregular sampling.
Continuous normalizing flows~\cite{grathwohl2019ffjord} further enable flexible density transport via learned vector fields.
Flow Matching~\cite{lipman2023flow} further offers an efficient objective for learning probability paths without simulation.
Building upon these advances, ImageFlowNet~\cite{liu2025imageflownet} learns multiscale image representations and evolves them using position-parameterized neural ODE/SDEs to forecast disease trajectories from baseline scans.
Trajectory Flow Matching (TFM)~\cite{zhang2024tfm} leverages flow matching to train neural SDEs in a simulation-free manner, improving scalability and stability for stochastic time series modeling.
IMMFM~\cite{islam2026longitudinal} further proposes interpolative multi-marginal flow matching to jointly learn stochastic dynamics consistent with multiple observed time points.
$\Delta$-LFM~\cite{chen2026learning} models disease progression as a velocity field in a semantically aligned latent space, enabling interpretable patient-specific trajectory generation. 
While these methods demonstrate strong performance in longitudinal forecasting, they typically learn progression end-to-end, without explicitly disentangling structural deformation from appearance variation, or incorporating domain-specific priors governing lesion growth.

\subsection{Physics-Based Modeling of Brain Lesions}

Physics-based models characterize lesion growth through mechanistic descriptions of underlying biological processes.
A common formulation relies on reaction–diffusion PDEs, which {model lesion cell density evolution via proliferation and diffusive spread}.
Grounded in classical population dynamics, these models have been widely used to describe tumor and lesion progression~\cite{tracqui1995mathematical,swanson2000quantitative,clatz2005realistic, raissi2019physics, tuncc2021modeling}. 
A canonical example is the Fisher-KPP equation, which combines diffusion and logistic growth to model wave-like lesion expansion~\cite{rockne2010predicting}.
Analytical and numerical studies of such dynamics provide insights into propagation speed, spatial spread, and parameter influence~\cite{murray2002mathematical, swanson2003virtual}.

A key challenge is the inverse problem: estimating patient-specific tumor cell distributions or model parameters from imaging data~\cite{hogea2008image,konukoglu2010extrapolating}.
Prior work commonly formulates tumor personalization as a PDE-constrained optimization problem, which requires repeated forward PDE solves and is therefore computationally expensive~\cite{mang2018pde,scheufele2020image}. 
To address scalability, recent works integrate deep learning into physics-based inference.
Learn-Morph-Infer~\cite{ezhov2023learn} proposes amortized inference to directly predict patient-specific tumor cell distributions from MRI, significantly reducing personalization time while remaining compatible with reaction-diffusion models.
GliODIL~\cite{balcerak2025individualizing} softly assimilates imaging data and Fisher-KPP constraints through a discrete optimization framework, improving individualized radiotherapy planning.
Similarly,~\cite{weidner2024learnable} incorporates a learnable deep-learning prior to initialize physics-based evolutionary sampling, accelerating convergence in inverse tumor modeling.
Physics-regularized multi-modal image assimilation~\cite{balcerak2024physics} further integrates PDE growth and biomechanical constraints as regularization terms within a learnable discrete framework. 
Despite improved efficiency and hybrid modeling strategies, these approaches primarily focus on parameter estimation or hidden cell reconstruction.
Their physics constraints are applied during inverse inference, rather than embedded within the forward generative trajectory for longitudinal image forecasting.

\begin{figure}[t]
\centering
\includegraphics[width=1\linewidth]{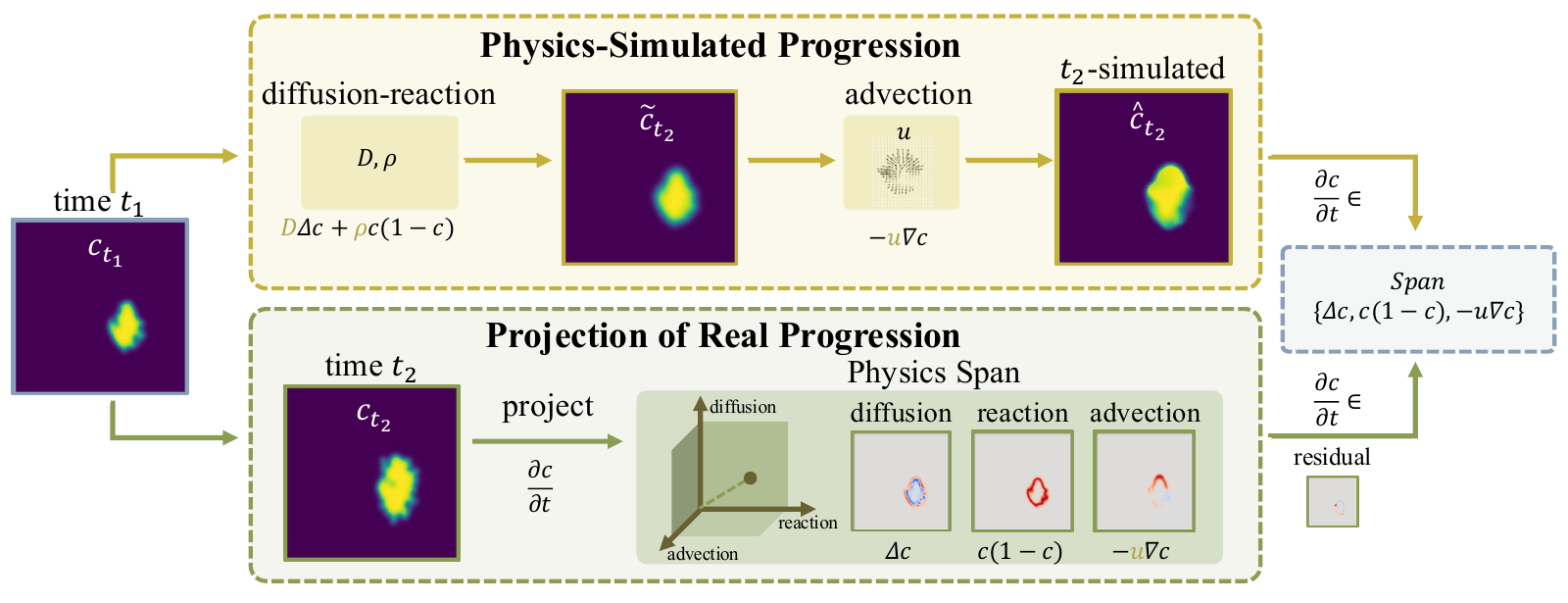}
\caption{\textbf{Physics-simulated lesion progression and projection of real progression onto the physics span.} 
\textbf{Top:} lesion cell concentration evolution simulated under the diffusion-reaction-advection formulation. 
\textbf{Bottom:} projection of real lesion concentration changes onto the same diffusion-reaction-advection span, illustrating how real disease progression can be approximated within a physics-grounded subspace.}
\label{fig:projection}
\end{figure}

\section{Method}

Sec.~\ref{subsec:physics_modeling} introduces our physics-grounded lesion evolution model, including PDE-based lesion growth dynamics (Sec.~\ref{subsubsec:pde_growth}) and PDE-regularized progression learning (Sec.~\ref{subsubsec:pde_progression}). Sec.~\ref{subsec:disentangled_fm} presents the disentangled flow matching framework, covering morphology evolution, intensity evolution, and the inference procedure.

\subsection{Physics-Grounded Lesion Evolution Modeling}
\label{subsec:physics_modeling}

\subsubsection{PDE-based Lesion Growth Dynamics\vspace{0.15cm}}
\label{subsubsec:pde_growth}

\hfill \break 
\noindent A commonly used formulation for modeling lesion evolution is the Fisher-Kolmogorov-Petrovsky-Piskunov (Fisher-KPP) type diffusion-reaction PDE, which describes the spatiotemporal dynamics of lesion cell concentration~\cite{metzcar2024review, yin2019review}. 

Let $c(\mathbf{x}, t) \in [0,1]$ denote the normalized lesion cell concentration at spatial location $\mathbf{x}$ and time $t$. 
We have the Fisher-KPP dynamics as
\begin{equation}
\frac{\partial c}{\partial t}
=
D \Delta c
+
\rho\, c(1-c)\,,
\label{eq:fisher_kpp}
\end{equation}
where $D=D(t)$ is a diffusion coefficient which controls the spatial infiltration, 
capturing wave-like cell expansion, random movement, and spreading;
$\rho=\rho(t)$ is the proliferation (i.e., reaction) rate governing local growth 
such as cell division and multiplication.  
To further account for the transport effects induced by larger structural deformation or mass effect~\cite{subramanian2020multiatlas}, \cref{eq:fisher_kpp} is often augmented with an advection term and becomes
\begin{equation}
\frac{\partial c}{\partial t}
=
D \Delta c
+
\rho\, c(1-c)
-
\mathbf{u}\!\cdot\!\nabla c\,,
\label{eq:rad}
\end{equation}
$\mathbf{u}=\mathbf{u}(\mathbf{x}, t)$ is a spatiotemporally varying advection field. 
These three components correspond to diffusion, reaction, and advection processes, respectively.

\subsubsection{PDE-Regularized Progression Learning\vspace{0.15cm}} 
\label{subsubsec:pde_progression}

\hfill \break
\noindent In practice, the underlying patient-specific PDE parameters $(D, \rho, \mathbf{u})$ in \cref{eq:rad} are unknown.
Instead of explicitly solving the full inverse problem, we design a physics-grounded formulation that embeds the PDE into learning.

\paragraph{\textbf{Physics-Simulated Progression.}}
Given two longitudinal observations, $c_{t_1}$ at $t_1$ and $c_{t_2}$ at $t_2$, we first simulate a progression based on \cref{eq:rad}, as illustrated in Fig.~\ref{fig:projection} (top).
We begin with the diffusion-reaction components and estimate the optimal $(D, \rho)$ via least squares fitting:
\begin{equation}
(D, \rho)
=
\arg\min_{D,\rho}
\left\|
\frac{c_{t_2}-c_{t_1}}{t_2 - t_1}
-
D \Delta c_{t_1}
-
\rho\, c_{t_1}(1-c_{t_1})
\right\|_2^2 \,.
\label{eq:ls_dr}
\end{equation}
Given the estimated $(D, \rho)$, and the corresponding diffusion-reaction prediction
$\tilde{c}_{t_2}$ from $c_{t_1}$, the remaining discrepancy between
$\tilde{c}_{t_2}$ and the observed $c_{t_2}$ is attributed to transport effects.

Then we approximate the residual motion for $\mathbf{u}$ in the advection term through image-based transport estimation.
Specifically, we estimate a dense displacement field 
$\mathbf{d}(\mathbf{x})$
that aligns $\tilde{c}_{t_2}$ to $c_{t_2}$:
\begin{equation}
\mathbf{d}
=
\arg\min_{\mathbf{d}}~
\mathcal{E}\big(
\tilde{c}_{t_2}(\mathbf{x}),
c_{t_2}(\mathbf{x} + \mathbf{d}(\mathbf{x}))
\big) \,,
\label{eq:optical_flow}
\end{equation}
where $\mathcal{E}(\cdot)$ denotes a variational optical-flow objective.
In practice, we employ a TV-L1 optical flow formulation~\cite{zach2007duality} to obtain $\mathbf{d}$. 
The advection field is then
\begin{equation}
\mathbf{u}(\mathbf{x})
=
-
\frac{\mathbf{d}(\mathbf{x})}{t_2 - t_1},
\label{eq:u_from_flow}
\end{equation}
which converts displacement into instantaneous transport velocity.
The negative sign follows from the semi-Lagrangian formulation used in forward simulation.

As shown in Fig.~\ref{fig:projection} (top), the full diffusion-reaction-advection in \cref{eq:rad} is obtained with
$(D, \rho, \mathbf{u})$ to form
a physics-driven progression that closely approximates
the observed longitudinal evolution.

\paragraph{\textbf{Projection of Real Progression onto Physics Span.}}
Given two observations $c_{t_1}$ and $c_{t_2}$, the evolution of lesion cell concentration is approximated as
\vspace{-0.1cm}
\begin{equation}
v_{\text{real}}
=
\frac{c_{t_2}-c_{t_1}}{t_2 - t_1} \,.
\label{eq:v_real}
\vspace{0.1cm}
\end{equation}
With the estimated fields $(D,\rho,\mathbf{u})$, 
and following the formulation in Eq.~\eqref{eq:rad}, $v_{\text{real}}$ can be approximated via a combination of diffusion-, reaction-, and advection-induced components, evaluated at $t_1$:
\begin{equation}
v_{\text{real}}(\mathbf{x})
\approx
D \, \Delta c_{t_1}(\mathbf{x})
+
\rho \, c_{t_1}(\mathbf{x})\big(1-c_{t_1}(\mathbf{x})\big)
-
\mathbf{u}(\mathbf{x})\cdot \nabla c_{t_1}(\mathbf{x}) \,.
\label{eq:v_decomp}
\end{equation}
Accordingly, we define the physics span formed by the diffusion, reaction, and advection components as
\begin{equation}
\mathcal{S}(c_{t_1};\mathbf{u})
=
\mathrm{span}
\left\{
\Delta c_{t_1},\;
c_{t_1}(1-c_{t_1}),\;
-\mathbf{u} \!\cdot\! \nabla c_{t_1}
\right\} \,.
\label{eq:pde_span_u}
\end{equation}

Therefore, projecting $v_{\text{real}}$ onto $\mathcal{S}(c_{t_1};\mathbf{u})$ reduces to solving a least-squares problem over the scalar coefficients $(\alpha,\beta,\gamma)$:
\begin{equation}
(\alpha,\beta,\gamma)
=
\arg\min_{\alpha,\beta,\gamma}
\left\|
v_{\text{real}}
-
\alpha\,\Delta c_{t_1}
-
\beta\,c_{t_1}(1-c_{t_1})
-
\gamma\,\big(-\mathbf{u}\!\cdot\!\nabla c_{t_1}\big)
\right\|_2^2 \,,
\label{eq:proj_ls}
\end{equation}
and the projected velocity is $\Pi_{\mathcal{S}}(v_{\text{real}})=
\alpha\,\Delta c_{t_1}+
\beta\,c_{t_1}(1-c_{t_1})+
\gamma\,(-\mathbf{u}\!\cdot\!\nabla c_{t_1})$.

To quantify the degree to which real longitudinal evolution conforms to this PDE-induced structure, we report the normalized projection error
\begin{equation}
\epsilon
=
\frac{\left\|v_{\text{real}}-\Pi_{\mathcal{S}}(v_{\text{real}})\right\|_2}
     {\left\|v_{\text{real}}\right\|_2} \,,
\label{eq:proj_error}
\end{equation}
which is visualized as a residual map in Fig.~\ref{fig:projection} (bottom).

\begin{figure}[t]
\centering
\includegraphics[width=1\linewidth]{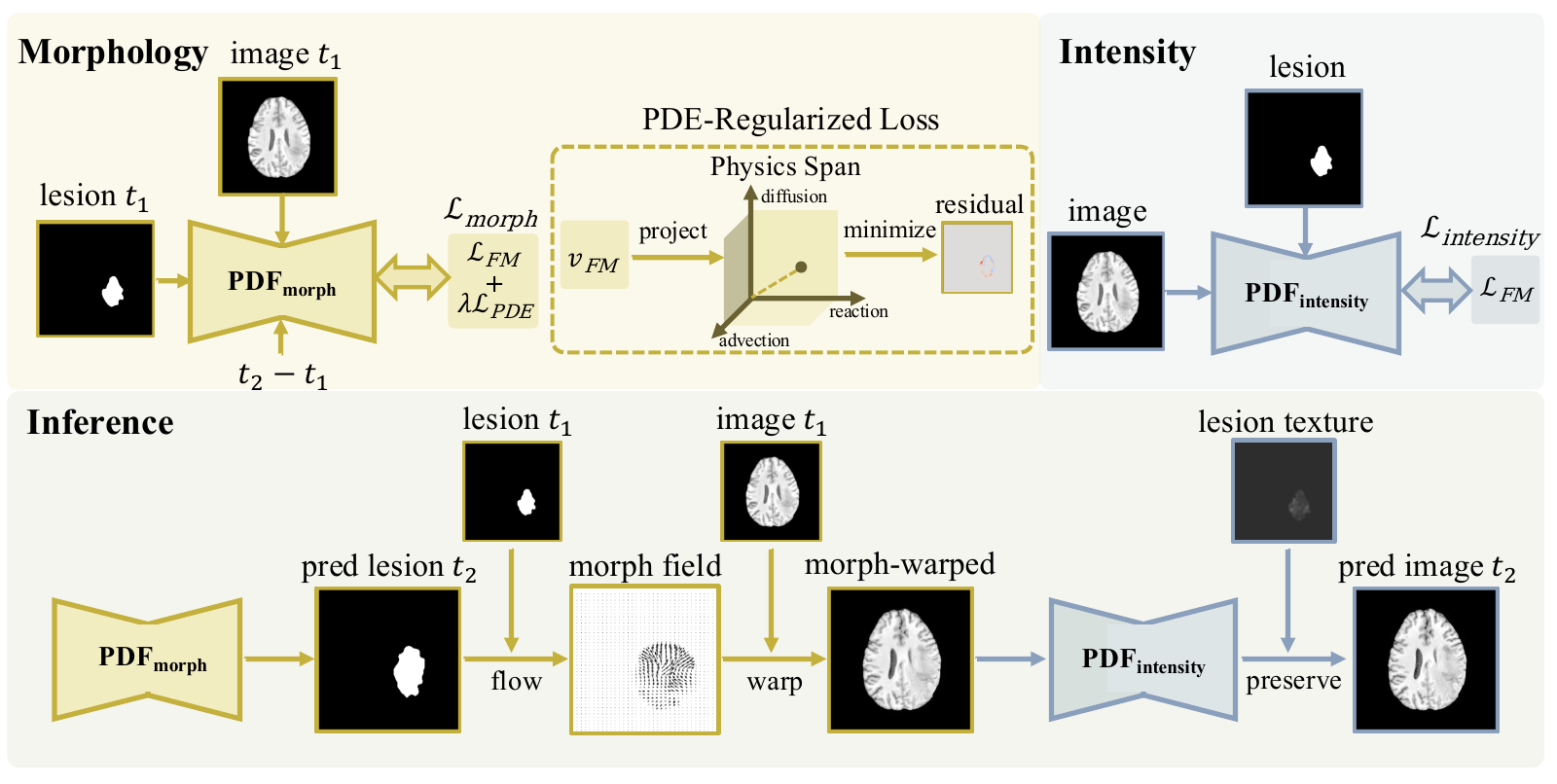}
\caption{\textbf{Overview of the proposed PDF framework.} 
\textbf{Top-left:} morphology flow matching network, which describes lesion growth and structural deformation. The proposed PDE-regularized loss minimizes the residual of the predicted velocity after projection onto the physics span. 
\textbf{Top-right:} intensity flow matching network, which models signal evolution conditioned on the image and lesion region. 
\textbf{Bottom:} inference procedure, where morphological evolution is first predicted to update lesion-induced structural changes, followed by intensity evolution to generate the final prediction.
}
\label{fig:overview}
\end{figure}

\paragraph{\textbf{PDE-Regularized Loss.}}
To incorporate this physics prior into learning, we constrain the predicted velocity field $v_\theta$ (i.e., evolution of lesion cell concentration in \cref{eq:v_real}) to remain close to the physics span $\mathcal{S}(c_{t_1};\mathbf{u})$. 
Specifically, for a given $v_\theta$, we estimate its projection onto $\mathcal{S}(c_{t_1};\mathbf{u})$ by solving
\begin{equation}
(\alpha,\beta,\gamma)
=
\arg\min_{\alpha,\beta,\gamma}
\left\|
v_\theta
-
\alpha\,\Delta c_{t_1}
-
\beta\,c_{t_1}(1-c_{t_1})
-
\gamma\,\big(-\mathbf{u}\!\cdot\!\nabla c_{t_1}\big)
\right\|_2^2 \,.
\label{eq:proj_theta}
\end{equation}
Denoting the projected velocity as
\begin{equation}
\Pi_{\mathcal{S}}(v_\theta)
=
\alpha\,\Delta c_{t_1}
+
\beta\,c_{t_1}(1-c_{t_1})
+
\gamma\,\big(-\mathbf{u}\!\cdot\!\nabla c_{t_1}\big) \,,
\end{equation}
we define the PDE regularization term as the squared projection residual:
\begin{equation}
\mathcal{L}_{\text{PDE}}
=
\left\|
v_\theta
-
\Pi_{\mathcal{S}}(v_\theta)
\right\|_2^2 \,.
\label{eq:pde_loss}
\end{equation}
This regularizer penalizes deviations of the estimated lesion cell concentration evolution from the diffusion-reaction-advection subspace, thereby embedding physics-grounded constraints into learning without explicitly solving a full PDE inverse problem.

\subsection{Disentangled Flow Matching Framework}
\label{subsec:disentangled_fm}

As illustrated in Fig.~\ref{fig:overview}, instead of directly predicting a future scan from a baseline image, we decompose longitudinal image prediction of lesion progression into two complementary components: 
\emph{morphology evolution}, capturing lesion growth and structural deformation, and 
\emph{intensity evolution}, modeling the overall image appearance changes conditioned on the predicted lesion evolution morphology.
Each component is modeled with a corresponding flow matching module.

\paragraph{\textbf{Morphology Flow Matching.}}
Let $M_{t_1}$ and $M_{t_2}$ denote the binary lesion masks at time $t_1$ and $t_2$, respectively, with time interval as $t_2 - t_1$.
To obtain a continuous field for PDE modeling, each binary mask is converted into a smooth signed distance function and mapped to a normalized lesion cell concentration $c_t(\mathbf{x}) \in [0,1]$, where larger values indicate higher lesion cell density.
The resulting concentrations $c_{t_1}$ and $c_{t_2}$ are then used to estimate the PDE parameters, $(D, \rho, \mathbf{u})$, in Sec.~\ref{subsubsec:pde_progression}.

We learn a velocity field that describes the morphological evolution associated with lesion growth,
\begin{equation}
v_\theta
=
f_\theta(M_{t_1};\, I_{t_1}, t_2 - t_1) \,,
\end{equation}
where $I_{t_1}$, the image at $t_1$, serves as a structural context, $t_2 - t_1$ encodes temporal conditioning.
The network $f_\theta$ predicts the instantaneous morphological velocity. 
In flow matching, the empirical velocity target is approximated by
\begin{equation}
v_{\text{morph}}
=
\frac{M_{t_2} - M_{t_1}}{t_2 - t_1} \,,
\end{equation} 
and the standard flow matching objective is defined as,
\begin{equation}
\mathcal{L}_{\text{FM}}
=
\left\|
v_\theta - v_{\text{morph}}
\right\|_2^2 \,.
\end{equation}
Incorporating the PDE regularization defined in Eq.~\eqref{eq:pde_loss} to enforce physics consistency, we have
\begin{equation}
\label{eq:morph_loss}
\mathcal{L}_{\text{morph}}
=
\mathcal{L}_{\text{FM}}
+
\lambda \, \mathcal{L}_{\text{PDE}} \,,
\end{equation}
where $\lambda$ controls the strength of the physics-grounded constraint.

\paragraph{\textbf{Intensity Flow Matching.}}

While morphology flow accounts for structural deformation, intensity evolution captures signal variations conditioned on lesion evolution.
During training, we adopt a flow-matching scheme in image space. 
Specifically, a noisy sample $z_\tau$ is generated along a predefined interpolation trajectory in image space, parameterized by $\tau \in [0,1]$. 
The intensity evolution velocity is defined as
\begin{equation}
v_\phi
=
g_\phi(z_\tau;\, M) \,,
\end{equation}

Following the standard flow matching formulation, the target velocity $v_{\text{intensity}}$ is defined by the chosen interpolation path in image space.
The image intensity learning objective is therefore
\begin{equation}
\mathcal{L}_{\text{intensity}}
=
\left\|
v_\phi - v_{\text{intensity}}
\right\|_2^2 \,.
\end{equation}

This formulation accounts for changes in image intensity, in addition to structural deformations resulting from lesion morphology.

\paragraph{\textbf{Inference.}}

At inference time, the predicted morphological velocity $v_\theta$ is first integrated over the time interval ($t_2 - t_1$) to obtain the evolved lesion mask $\hat{M}_{t_2}$. 
To obtain the dense structural deformation, we estimate a displacement field ($\mathbf{f}$) between $\hat{M}_{t_2}$ and $M_{t_1}$ with a variational optical flow formulation.
Specifically, 
\begin{equation}
\mathbf{f}(\mathbf{x})
=
\arg\min_{\mathbf{f}}
\mathcal{E}\big(
\hat{M}_{t_2}(\mathbf{x}),
M_{t_1}(\mathbf{x}+\mathbf{f}(\mathbf{x}))
\big)\,,
\label{eq:flow_estimation}
\end{equation}
where $\mathcal{E}$ denotes a TV-L1 optical flow objective.

Given the estimated dense morph field $\mathbf{f}(\mathbf{x})$, the baseline image $I_{t_1}$ at time $t_1$ is warped using a semi-Lagrangian formulation:
\begin{equation}
\hat{I}^{\text{morph}}(\mathbf{x})
=
I_{t_1}\!\left(
\mathbf{x}
+
\mathbf{f}(\mathbf{x})
\right) \,,
\label{eq:warp}
\end{equation}
where bilinear interpolation is used for sampling, and the sampling coordinates are clipped to remain within structural boundaries.
This step models structural deformation induced by lesion growth.

While $\hat{I}^{\text{morph}}$ captures structural displacement, it does not account for the image intensity changes induced by the estimated lesion cell concentration variations.
To model appearance evolution, we apply the intensity flow matching module to the warped image, $\hat{I}^{\text{morph}}$. 
Specifically, $\hat{I}^{\text{morph}}$ serves as the structural initialization, and is refined by integrating the learned intensity velocity field conditioned on the baseline lesion mask $M_{t_1}$.
To preserve lesion-specific texture patterns, high-frequency components extracted from $I_{t_1}$ are incorporated as additional guidance during refinement.

The final prediction $\hat{I}_{t_2}$ is obtained after intensity flow integration, producing appearance changes that are consistent with both the predicted morphology and the intrinsic lesion texture.

\section{Experiments}
\subsection{Experimental Settings}

\paragraph{\textbf{Datasets.}}
We evaluate our framework on three publicly available longitudinal brain disease cohorts, covering diverse modalities and pathology types. In each dataset, images are longitudinally co-registered per subject during preprocessing.

\vspace{0.15cm}
\noindent \textbf{- UCSF.}
The UCSF-ALPTDG dataset \cite{fields2024university} provides longitudinal MRI scans of 302 adult patients with \textit{brain glioma tumors}. Each patient has two time points with expert-annotated lesion masks for precise evaluation of tumor progression. In this work, we use the T1-weighted MRI modality for longitudinal modeling.

\vspace{0.15cm}
\noindent \textbf{- LUMIERE.}
The LUMIERE dataset \cite{suter2022lumiere} contains longitudinal MRI scans of 91 patients with \textit{glioblastoma tumors}, along with corresponding lesion masks. 
Each patient underwent one pre-operative scan up to five years, resulting in 795 longitudinal series with 2-18 time points per case.
Following prior work, we retain only post-operative scans to model natural tissue evolution after surgery, and use contrast-enhanced T1-weighted MRI scans in our experiments.

\vspace{0.15cm}
\noindent \textbf{- LMSLS.}
The LMSLS dataset \cite{carass2017longitudinal} contains longitudinal FLAIR MRI scans from 79 data points with \textit{multiple sclerosis (MS) lesions}. Each subject has an average of 4.4 visits collected over approximately five years with expert-annotated lesion masks. We adopt FLAIR-weighted MRI for longitudinal lesion modeling.

\paragraph{\textbf{Lesion Cell Concentration Map Construction.}}

For each binary lesion mask $M$, we first convert it into a continuous cell concentration map, via a signed distance function (SDF) that measures the distance from each pixel to the lesion boundary ($\partial M$). The signed distance function $d(\mathbf{x})$ is defined as
\begin{equation}
d(\mathbf{x}) =
\begin{cases}
~-\min_{\mathbf{y}\in \partial M} |\mathbf{x}-\mathbf{y}|, & \mathbf{x} \in M, \\
~\min_{\mathbf{y}\in \partial M} |\mathbf{x}-\mathbf{y}|, & \mathbf{x} \notin M ,
\end{cases}
\end{equation}
where negative (positive) values indicate locations inside (outside) the lesion.
We further normalize $d(\mathbf{x})$ with sigmoid transformation
$
c(\mathbf{x}) = 1 / \left(1 + \exp(\frac{d(\mathbf{x})}{\tau})\right),
$
where $\tau$ controls the smoothness of the transition around the lesion boundary.
This transformation produces a continuous lesion cell concentration field $c(\mathbf{x}) \in [0,1]$, where larger values correspond to higher lesion cell density.

\paragraph{\textbf{Evaluation Metrics.}}
We evaluate longitudinal prediction performance from both image fidelity and disease region consistency perspectives. 
For lesion-level assessment, we compute the Dice-S{\o}rensen coefficient (DSC) and Hausdorff distance (HD) between predicted and ground-truth lesion masks. 
DSC measures volumetric overlap between regions, while HD quantifies the maximum boundary discrepancy. 
Lesion regions are obtained using dataset-specific image segmentation networks following~\cite{liu2025imageflownet}.
For image-level assessment, we report average L1 error, peak signal-to-noise ratio (PSNR) and structural similarity index (SSIM) for reconstruction quality and structural consistency of the predicted images.

\paragraph{\textbf{Implementation Details.}}
All time points in each longitudinal series were rigidly registered to ensure spatial alignment across visits. 
We partition the data into training, validation, and test sets using a 6:2:2 ratio. 
To prevent data leakage, splitting was performed at the longitudinal series level, ensuring that all scans from the same patient were assigned to the same subset.
All flow matching models are implemented using a U-Net backbone with a shared encoder-decoder structure. 
Experiments were conducted on an NVIDIA RTX PRO 6000 GPU with CUDA version 13.0 and PyTorch 2.9.1.

\begin{figure}[t]
\centering
\fontsize{6.5}{10}\selectfont
\setlength{\tabcolsep}{1pt}
\renewcommand{\arraystretch}{1.45}

\begin{tabular}{c@{\hspace{4pt}}cc@{\hspace{5pt}}ccccc}

& {\ttfamily\bfseries Ground-truth ($t_2$)}
& {\ttfamily\bfseries Input ($t_1$)}
& {\ttfamily\bfseries T-UNet}
& {\ttfamily\bfseries I$^2$SB}
& {\ttfamily\bfseries TFM}
& {\ttfamily\bfseries ImageFlowNet}
& {\ttfamily\bfseries PDF (Ours)} \\[1pt]

\makebox[5pt][r]{\raisebox{-0.5\height}
{\texttt{\shortstack{Image\\(Tumor)}}}}%
& \begin{minipage}{0.12\linewidth}
  \centering
  {\fontsize{5}{10}\selectfont \hspace*{-0.04cm}PSNR$\uparrow$/SSIM$\uparrow$}\\[-0.pt]
  \includegraphics[width=\linewidth]{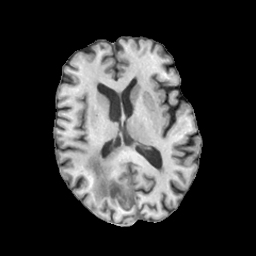}
  \end{minipage}
& \begin{minipage}{0.12\linewidth}
  \centering
  {\fontsize{5}{10}\selectfont 28.4/0.88}\\[-0.pt]
  \includegraphics[width=\linewidth]{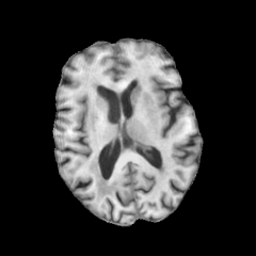}
  \end{minipage}
& \begin{minipage}{0.12\linewidth}
  \centering
  {\fontsize{5}{10}\selectfont 28.1/0.88}\\[-0.pt]
  \includegraphics[width=\linewidth]{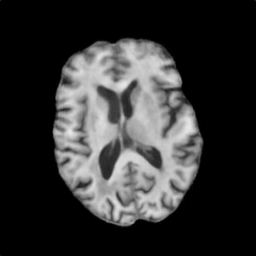}
  \end{minipage}
& \begin{minipage}{0.12\linewidth}
  \centering
  {\fontsize{5}{10}\selectfont 28.4/0.88}\\[-0.pt]
  \includegraphics[width=\linewidth]{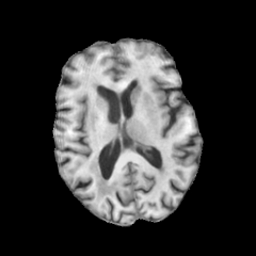}
  \end{minipage}
& \begin{minipage}{0.12\linewidth}
  \centering
  {\fontsize{5}{10}\selectfont 28.4/0.88}\\[-0.pt]
  \includegraphics[width=\linewidth]{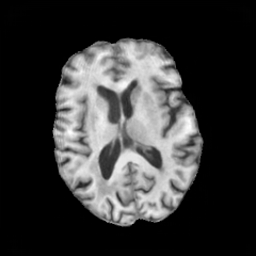}
  \end{minipage}
& \begin{minipage}{0.12\linewidth}
  \centering
  {\fontsize{5}{10}\selectfont 26.4/0.81}\\[-0.pt]
  \includegraphics[width=\linewidth]{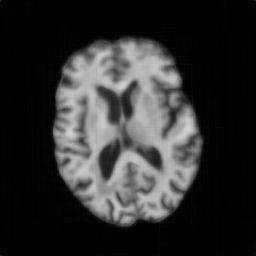}
  \end{minipage}
& \begin{minipage}{0.12\linewidth}
  \centering
  {\fontsize{5}{10}\selectfont {28.8/0.89}}\\[-0.pt]
  \includegraphics[width=\linewidth]{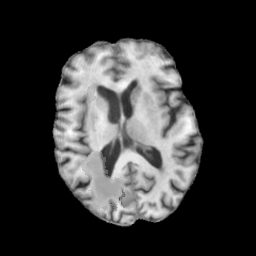}
  \end{minipage}
\\[21pt]

\makebox[5pt][r]{\raisebox{-0.5\height}{\texttt{\shortstack{w/\\Lesion\\Region}}}}
& \begin{minipage}{0.12\linewidth}
  \centering
  {\fontsize{5}{10}\selectfont DSC$\uparrow$/HD$\downarrow$}\\[-0.pt]
  \includegraphics[width=\linewidth]{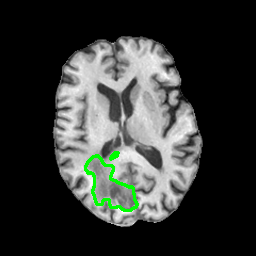}
  \end{minipage}
& \begin{minipage}{0.12\linewidth}
  \centering
  {\fontsize{5}{10}\selectfont 0.53/24.2}\\[-0.pt]
  \includegraphics[width=\linewidth]{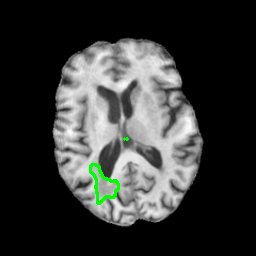}
  \end{minipage}
& \begin{minipage}{0.12\linewidth}
  \centering
  {\fontsize{5}{10}\selectfont 0.53/20.4}\\[-0.pt]
  \includegraphics[width=\linewidth]{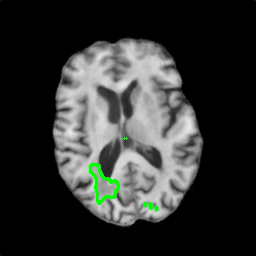}
  \end{minipage}
& \begin{minipage}{0.12\linewidth}
  \centering
  {\fontsize{5}{10}\selectfont 0.54/24.2}\\[-0.pt]
  \includegraphics[width=\linewidth]{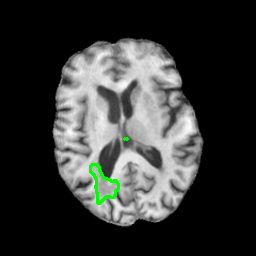}
  \end{minipage}
& \begin{minipage}{0.12\linewidth}
  \centering
  {\fontsize{5}{10}\selectfont 0.54/24.0}\\[-0.pt]
  \includegraphics[width=\linewidth]{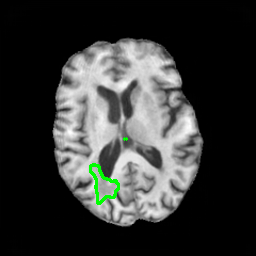}
  \end{minipage}
& \begin{minipage}{0.12\linewidth}
  \centering
  {\fontsize{5}{10}\selectfont 0.48/24.6}\\[-0.pt]
  \includegraphics[width=\linewidth]{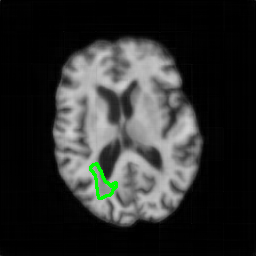}
  \end{minipage}
& \begin{minipage}{0.12\linewidth}
  \centering
  {\fontsize{5}{10}\selectfont {0.74/18.2}}\\[-0.pt]
  \includegraphics[width=\linewidth]{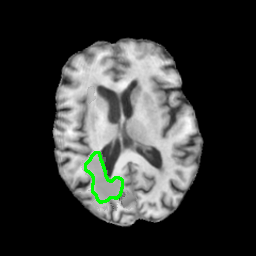}
  \end{minipage}
\\[21pt]

\makebox[5pt][r]{\raisebox{-0.5\height}
{\texttt{\shortstack{Image\\(Tumor)}}}}
& \begin{minipage}{0.12\linewidth}
  \centering
  {\fontsize{5}{10}\selectfont \hspace*{-0.04cm}PSNR$\uparrow$/SSIM$\uparrow$}\\[-0.pt]
  \includegraphics[width=\linewidth]{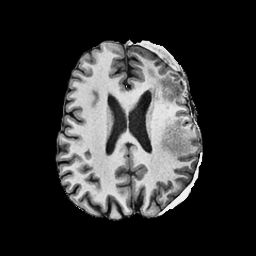}
  \end{minipage}
& \begin{minipage}{0.12\linewidth}
  \centering
  {\fontsize{5}{10}\selectfont 28.4/0.88}\\[-0.pt]
  \includegraphics[width=\linewidth]{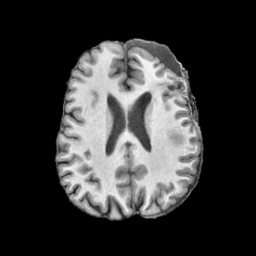}
  \end{minipage}
& \begin{minipage}{0.12\linewidth}
  \centering
  {\fontsize{5}{10}\selectfont 27.1/0.89}\\[-0.pt]
  \includegraphics[width=\linewidth]{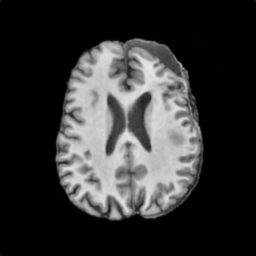}
  \end{minipage}
& \begin{minipage}{0.12\linewidth}
  \centering
  {\fontsize{5}{10}\selectfont 26.4/0.89}\\[-0.pt]
  \includegraphics[width=\linewidth]{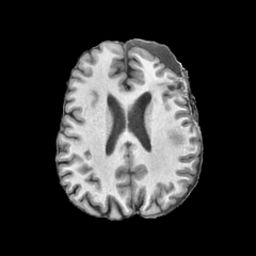}
  \end{minipage}
& \begin{minipage}{0.12\linewidth}
  \centering
  {\fontsize{5}{10}\selectfont 26.4/0.88}\\[-0.pt]
  \includegraphics[width=\linewidth]{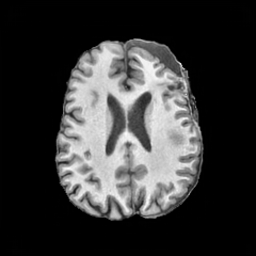}
  \end{minipage}
& \begin{minipage}{0.12\linewidth}
  \centering
  {\fontsize{5}{10}\selectfont 26.9/0.75}\\[-0.pt]
  \includegraphics[width=\linewidth]{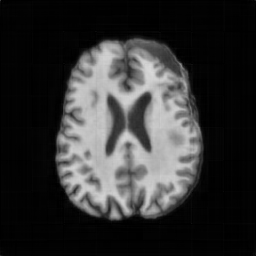}
  \end{minipage}
& \begin{minipage}{0.12\linewidth}
  \centering
  {\fontsize{5}{10}\selectfont {26.8/0.88}}\\[-0.pt]
  \includegraphics[width=\linewidth]{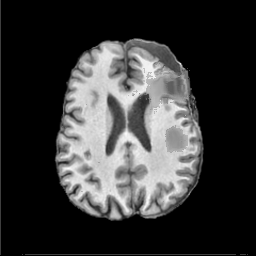}
  \end{minipage}
\\[21pt]

\makebox[5pt][r]{\raisebox{-0.5\height}{\texttt{\shortstack{w/\\Lesion\\Region}}}}
& \begin{minipage}{0.12\linewidth}
  \centering
  {\fontsize{5}{10}\selectfont DSC$\uparrow$/HD$\downarrow$}\\[-0.pt]
  \includegraphics[width=\linewidth]{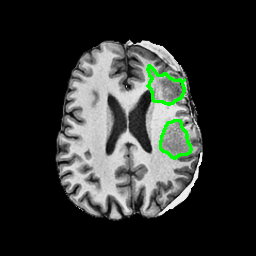}
  \end{minipage}
& \begin{minipage}{0.12\linewidth}
  \centering
  {\fontsize{5}{10}\selectfont 0.52/21.6}\\[-0.pt]
  \includegraphics[width=\linewidth]{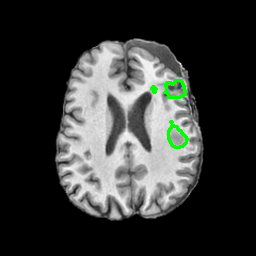}
  \end{minipage}
& \begin{minipage}{0.12\linewidth}
  \centering
  {\fontsize{5}{10}\selectfont 0.52/21.6}\\[-0.pt]
  \includegraphics[width=\linewidth]{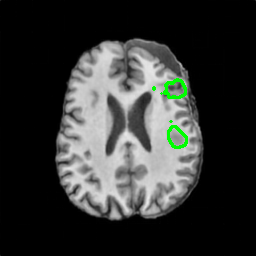}
  \end{minipage}
& \begin{minipage}{0.12\linewidth}
  \centering
  {\fontsize{5}{10}\selectfont 0.51/21.6}\\[-0.pt]
  \includegraphics[width=\linewidth]{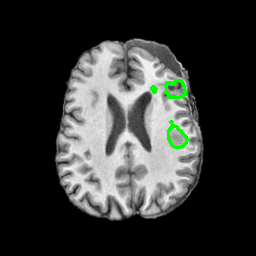}
  \end{minipage}
& \begin{minipage}{0.12\linewidth}
  \centering
  {\fontsize{5}{10}\selectfont 0.51/21.6}\\[-0.pt]
  \includegraphics[width=\linewidth]{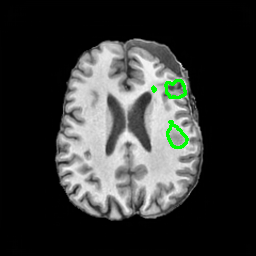}
  \end{minipage}
& \begin{minipage}{0.12\linewidth}
  \centering
  {\fontsize{5}{10}\selectfont 0.32/19.7}\\[-0.pt]
  \includegraphics[width=\linewidth]{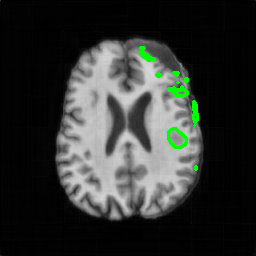}
  \end{minipage}
& \begin{minipage}{0.12\linewidth}
  \centering
  {\fontsize{5}{10}\selectfont {0.72/15.6}}\\[-0.pt]
  \includegraphics[width=\linewidth]{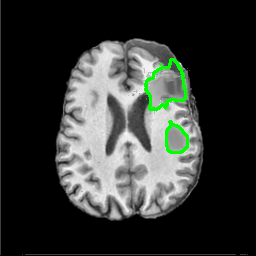}
  \end{minipage}
\\[22pt]

\makebox[5pt][r]{\raisebox{-0.5\height}
{\texttt{\shortstack{Image\\(MS)}}}}
& \begin{minipage}{0.12\linewidth}
  \centering
  {\fontsize{5}{10}\selectfont \hspace*{-0.04cm}PSNR$\uparrow$/SSIM$\uparrow$}\\[-7.8pt]
  \includegraphics[width=\linewidth, angle=270]{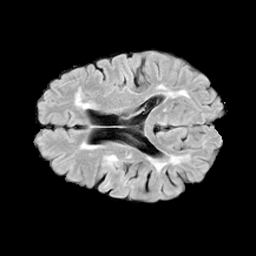}
  \end{minipage}
& \begin{minipage}{0.12\linewidth}
  \centering
  {\fontsize{5}{10}\selectfont 31.6/0.91}\\[-7.8pt]
\includegraphics[width=\linewidth, angle=270]{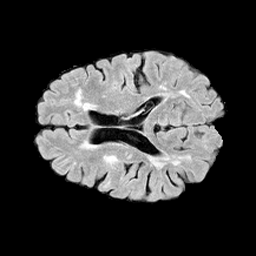}
  \end{minipage}
& \begin{minipage}{0.12\linewidth}
  \centering
  {\fontsize{5}{10}\selectfont 31.9/0.92}\\[-7.8pt]
  \includegraphics[width=\linewidth, angle=270]{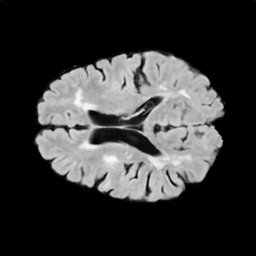}
  \end{minipage}
& \begin{minipage}{0.12\linewidth}
  \centering
  {\fontsize{5}{10}\selectfont 31.6/0.91}\\[-7.8pt]
  \includegraphics[width=\linewidth, angle=270]{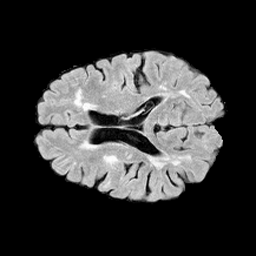}
  \end{minipage}
& \begin{minipage}{0.12\linewidth}
  \centering
  {\fontsize{5}{10}\selectfont 31.6/0.91}\\[-7.8pt]
  \includegraphics[width=\linewidth, angle=270]{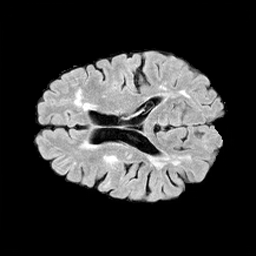}
  \end{minipage}
& \begin{minipage}{0.12\linewidth}
  \centering
  {\fontsize{5}{10}\selectfont 31.5/0.90}\\[-7.8pt]
  \includegraphics[width=\linewidth, angle=270]{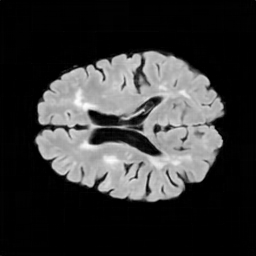}
  \end{minipage}
& \begin{minipage}{0.12\linewidth}
  \centering
  {\fontsize{5}{10}\selectfont {31.6/0.92}}\\[-7.8pt]
  \includegraphics[width=\linewidth, angle=270]{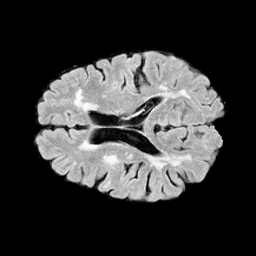}
  \end{minipage}
\\[22pt]

\makebox[5pt][r]{\raisebox{-0.5\height}{\texttt{\shortstack{w/\\Lesion\\Region}}}}
& \begin{minipage}{0.12\linewidth}
  \centering
  {\fontsize{5}{10}\selectfont DSC$\uparrow$/HD$\downarrow$}\\[-7.8pt]
  \includegraphics[width=\linewidth, angle=270]{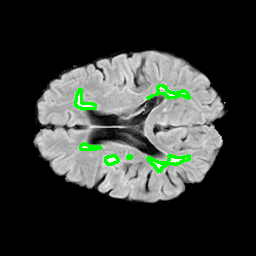}
  \end{minipage}
& \begin{minipage}{0.12\linewidth}
  \centering
  {\fontsize{5}{10}\selectfont 0.75/11.4}\\[-7.8pt]
\includegraphics[width=\linewidth, angle=270]{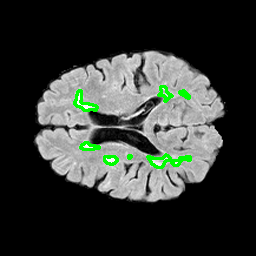}
  \end{minipage}
& \begin{minipage}{0.12\linewidth}
  \centering
  {\fontsize{5}{10}\selectfont 0.78/12.7}\\[-7.8pt]
  \includegraphics[width=\linewidth, angle=270]{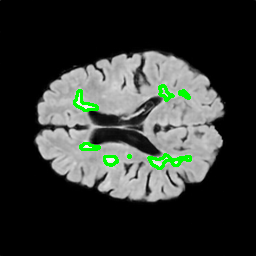}
  \end{minipage}
& \begin{minipage}{0.12\linewidth}
  \centering
  {\fontsize{5}{10}\selectfont 0.77/11.4}\\[-7.8pt]
  \includegraphics[width=\linewidth, angle=270]{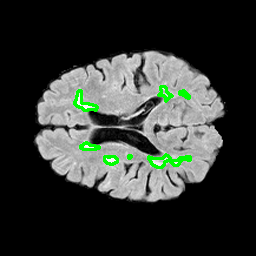}
  \end{minipage}
& \begin{minipage}{0.12\linewidth}
  \centering
  {\fontsize{5}{10}\selectfont 0.75/11.4}\\[-7.8pt]
  \includegraphics[width=\linewidth, angle=270]{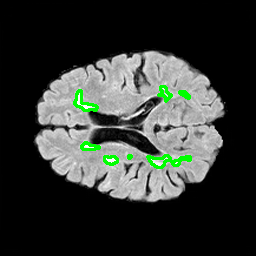}
  \end{minipage}
& \begin{minipage}{0.12\linewidth}
  \centering
  {\fontsize{5}{10}\selectfont 0.74/11.5}\\[-7.8pt]
  \includegraphics[width=\linewidth, angle=270]{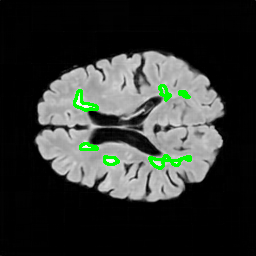}
  \end{minipage}
& \begin{minipage}{0.12\linewidth}
  \centering
  {\fontsize{5}{10}\selectfont {0.82/9.49}}\\[-7.8pt]
  \includegraphics[width=\linewidth, angle=270]{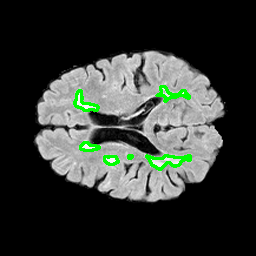}
  \end{minipage}

\end{tabular}

\caption{\textbf{Comparison of longitudinal modeling methods on two tumor cases and one MS lesion case.} 
Each case is presented in two rows: the first row shows the predicted images with PSNR/SSIM, the second row shows images overlaid with corresponding lesion regions, along with DSC/HD. 
Compared to the competing methods, \textbf{PDF} produces more accurate lesion boundaries and improved structural consistency.}
\label{fig:comparison}
\end{figure}


\newcommand{\methodcell}[1]{\raisebox{0.7ex}{#1}}

\newcommand{\mystackstrut}{\rule{0pt}{2.4ex}}
\newcommand{\mydepthstrut}{\rule[-1.0ex]{0pt}{0pt}}

\newcommand{\valstd}[2]{%
\shortstack[c]{\mystackstrut $#1$\\{\scriptsize$\pm#2$}\mydepthstrut}}

\newcommand{\bvalstd}[2]{%
\shortstack[c]{\mystackstrut $\mathbf{#1}$\\{\scriptsize$\pm#2$}\mydepthstrut}}

\begin{table*}[t]
\centering
\setlength{\tabcolsep}{0.85mm}
\renewcommand{\arraystretch}{1.3}
\fontsize{9}{10.4}\selectfont
\resizebox{1.\textwidth}{!}{%
\begin{tabular}{l ccccc@{\hskip 1.5mm} ccccc@{\hskip 1.5mm} ccccc}
\toprule
\multirow{2}{*}{\textbf{Method}}
& \multicolumn{5}{c}{\textbf{UCSF (Tumor)}}
& \multicolumn{5}{c}{\textbf{LUMIERE (Tumor)}}
& \multicolumn{5}{c}{\textbf{LMSLS (MS)}} \\
\cmidrule(lr){2-6}
\cmidrule(lr){7-11}
\cmidrule(lr){12-16}
& \textbf{DSC}$\uparrow$ 
& \textbf{HD}$\downarrow$ 
& \textbf{PSNR}$\uparrow$ 
& \textbf{SSIM}$\uparrow$ 
& \textbf{L1}$\downarrow$
& \textbf{DSC}$\uparrow$ 
& \textbf{HD}$\downarrow$ 
& \textbf{PSNR}$\uparrow$ 
& \textbf{SSIM}$\uparrow$ 
& \textbf{L1}$\downarrow$
& \textbf{DSC}$\uparrow$ 
& \textbf{HD}$\downarrow$ 
& \textbf{PSNR}$\uparrow$ 
& \textbf{SSIM}$\uparrow$ 
& \textbf{L1}$\downarrow$ \\
\midrule  

\methodcell{T-UNet~\cite{ronneberger2015u}}
& \valstd{0.728}{0.004}
& \valstd{27.322}{0.812}
& \bvalstd{31.739}{0.103}
& \valstd{0.920}{0.002}
& \valstd{0.085}{0.002}
& \valstd{0.520}{0.020}
& \valstd{38.135}{1.925}
& \valstd{36.753}{0.185}
& \valstd{0.845}{0.004}
& \valstd{0.097}{0.003}
& \valstd{0.673}{0.005}
& \valstd{26.354}{0.932}
& \valstd{29.248}{0.118}
& \valstd{0.805}{0.003}
& \valstd{0.083}{0.002}   \\

\methodcell{I$^2$SB~\cite{liu2023i2sb}} 
& \valstd{0.725}{0.005}
& \valstd{28.839}{0.941}
& \valstd{31.393}{0.097}
& \valstd{0.920}{0.002}
& \valstd{0.093}{0.002}
& \valstd{0.522}{0.035}
& \valstd{36.105}{1.720}
& \valstd{37.206}{0.348}
& \valstd{0.953}{0.003}
& \valstd{0.084}{0.003}
& \valstd{0.682}{0.006}
& \valstd{23.617}{0.784}
& \valstd{29.312}{0.094}
& \valstd{0.893}{0.002}
& \valstd{0.068}{0.003} \\

\methodcell{TFM~\cite{zhang2024tfm}}  
& \valstd{0.696}{0.006}
& \valstd{31.701}{1.284}
& \valstd{30.659}{0.125}
& \valstd{0.898}{0.003}
& \valstd{0.091}{0.002}
& \valstd{0.492}{0.015}
& \valstd{39.211}{1.530}
& \valstd{36.317}{0.406}
& \valstd{0.894}{0.001}
& \valstd{0.088}{0.002}
& \valstd{0.670}{0.007}
& \valstd{24.438}{0.851}
& \bvalstd{29.441}{0.102}
& \valstd{0.889}{0.003}
& \valstd{0.062}{0.004} \\

\methodcell{ImageFlowNet~\cite{liu2025imageflownet}} 
& \valstd{0.724}{0.004}
& \valstd{28.402}{0.893}
& \valstd{31.675}{0.108}
& \valstd{0.909}{0.003}
& \valstd{0.085}{0.002}
& \valstd{0.493}{0.033}
& \valstd{41.994}{1.653}
& \valstd{37.178}{0.346}
& \valstd{0.916}{0.003}
& \valstd{0.078}{0.003}
& \valstd{0.687}{0.005}
& \valstd{24.296}{0.741}
& \valstd{28.972}{0.113}
& \valstd{0.890}{0.003}
& \valstd{0.072}{0.002} \\

\methodcell{\textbf{PDF (Ours)}} 
& \bvalstd{0.752}{0.003}
& \bvalstd{25.952}{0.654}
& \valstd{31.364}{0.092}
& \bvalstd{0.920}{0.001}
& \bvalstd{0.072}{0.002}
& \bvalstd{0.534}{0.013}
& \bvalstd{33.672}{1.229}
& \bvalstd{37.216}{0.349}
& \bvalstd{0.957}{0.002}
& \bvalstd{0.071}{0.002}
& \bvalstd{0.696}{0.004}
& \bvalstd{20.867}{0.623}
& \valstd{29.314}{0.087}
& \bvalstd{0.900}{0.002}
& \bvalstd{0.058}{0.002} \\

\bottomrule
\end{tabular}%
}

\vspace{0.12cm}
\caption{\textbf{Quantitative comparison across three longitudinal datasets.}}
\label{tab:comparison}
\end{table*}

\subsection{Performance Comparison}
We compare our method with four representative longitudinal disease modeling approaches across all datasets: Time-conditional UNet (T-UNet)~\cite{ronneberger2015u}, I$^2$SB~\cite{liu2023i2sb}, Trajectory Flow Matching (TFM)~\cite{zhang2024tfm}, ImageFlowNet~\cite{liu2025imageflownet}. 
T-UNet integrates temporal information by injecting time embeddings into intermediate UNet representations. 
I$^2$SB is a diffusion-based image-to-image Schr\"odinger Bridge model. 
TFM formulates temporal evolution as flow matching in image space. 
ImageFlowNet models longitudinal dynamics via neural-ODE-based flow prediction.

Tab.~\ref{tab:comparison} reports quantitative results on UCSF, LUMIERE, and LMSLS.
On UCSF, our method achieves the highest DSC of 0.752 and the lowest HD of 25.952, outperforming T-UNet with a DSC of 0.728 and an HD of 27.322. 
Although the PSNR of our method is 31.364, slightly lower than 31.739 from T-UNet, the improved lesion-level metrics indicate more accurate boundary alignment.
On the LUMIERE dataset, our method achieves the best DSC of 0.534 and the lowest HD of 33.672, surpassing both T-UNet and I$^2$SB in segmentation accuracy. 
PSNR and SSIM are 37.216 and 0.957, respectively, demonstrating stable intensity reconstruction under post-operative tissue evolution.
On LMSLS, which involves subtle and gradual lesion progression, our method achieves the highest DSC of 0.696 and the lowest HD of 20.867, compared with 0.673 DSC and 26.354 HD from T-UNet. 
These results suggest improved sensitivity to small-scale lesion evolution.

Fig.~\ref{fig:comparison} presents qualitative comparisons.
The first two rows show a glioma case with a notable tumor region and structural deformation. 
Compared to the ground-truth follow-up, the initial input yields a DSC of 0.53 and an HD of 24.2, while our method achieves 0.74 and 18.2 and produces boundaries that closely match the ground truth. 
All competing models tend to underestimate the lesion morphological changes or generate less coherent lesion expansion. This demonstrates the advantage of our \textbf{\textit{Morphology Evolution}} module.
The third and fourth rows present a glioma case with prominent intensity changes. 
Our method achieves a DSC of 0.72 and an HD of 15.6, compared with 0.52 and 21.6 from the input, while better preserving appearance consistency. 
This highlights the contribution of our \textbf{\textit{Intensity Evolution}} module.
The last two rows illustrate an MS case, with subtle lesion progression. Our model successfully captures the lesion evolution in the bottom-right region of the image, while competing methods struggle to do so.
Overall, the qualitative results are consistent with the quantitative improvements observed across diverse disease dynamics.

\begin{figure}[t]
\centering

\resizebox{\textwidth}{!}{
\begin{tikzpicture}

\pgfmathsetmacro{\sx}{0}
\pgfmathsetmacro{\sy}{0}
\pgfmathsetmacro{\dx}{2.3}

\pgfmathsetmacro{\titley}{1.45}
\pgfmathsetmacro{\metricA}{-1.90}
\pgfmathsetmacro{\metricB}{-1.45}


\node at (\sx-\dx,\titley){\textbf{\texttt{Ground-truth} ($t_2$)}};
\node at (\sx,\titley){\textbf{\texttt{Input} ($t_1$)}};
\node at (\sx+\dx,\titley){\textbf{\texttt{T-UNet}}};
\node at (\sx+2*\dx,\titley){\textbf{\texttt{I$^2$SB}}};
\node at (\sx+3*\dx,\titley){\textbf{\texttt{TFM}}};
\node at (\sx+4*\dx,\titley){\textbf{\texttt{ImageFlowNet}}};
\node at (\sx+5*\dx,\titley){\textbf{\texttt{PDF (Ours)}}};


\node at (\sx-\dx,\sy)
{\includegraphics[width=0.18\linewidth]{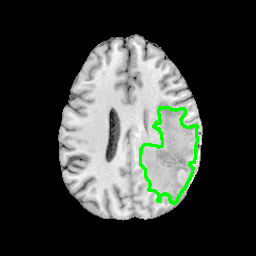}};

\draw[->,eccvblue!100,line width=0.8mm]
(\sx-\dx-0.2,\sy+0.8)--(\sx-\dx+0.13,\sy+0.2);


\node at (\sx,\sy)
{\includegraphics[width=0.18\linewidth]{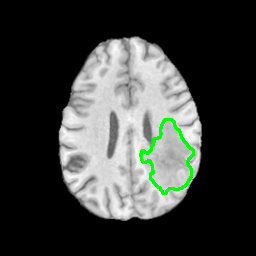}};

\draw[->,green,line width=0.8mm]
(\sx-0.2,\sy+0.8)--(\sx+0.13,\sy+0.2);


\node at (\sx+\dx,\sy)
{\includegraphics[width=0.18\linewidth]{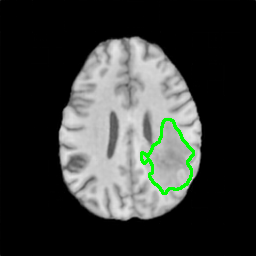}};

\draw[->,red,line width=0.8mm]
(\sx+\dx-0.2,\sy+0.8)--(\sx+\dx+0.13,\sy+0.2);


\node at (\sx+2*\dx,\sy)
{\includegraphics[width=0.18\linewidth]{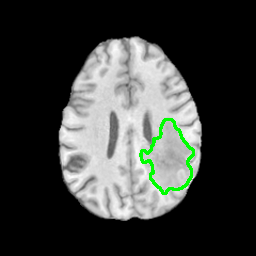}};

\draw[->,red,line width=0.8mm]
(\sx+2*\dx-0.2,\sy+0.8)--(\sx+2*\dx+0.13,\sy+0.2);


\node at (\sx+3*\dx,\sy)
{\includegraphics[width=0.18\linewidth]{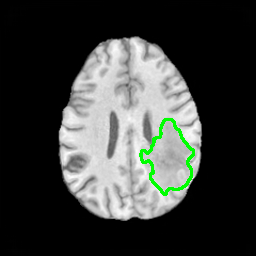}};

\draw[->,red,line width=0.8mm]
(\sx+3*\dx-0.2,\sy+0.8)--(\sx+3*\dx+0.13,\sy+0.2);


\node at (\sx+4*\dx,\sy)
{\includegraphics[width=0.18\linewidth]{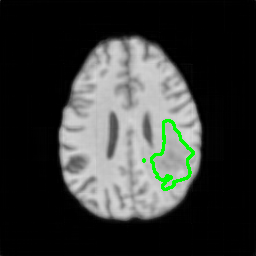}};

\draw[->,red,line width=0.8mm]
(\sx+4*\dx-0.2,\sy+0.8)--(\sx+4*\dx+0.13,\sy+0.2);


\node at (\sx+5*\dx,\sy)
{\includegraphics[width=0.18\linewidth]{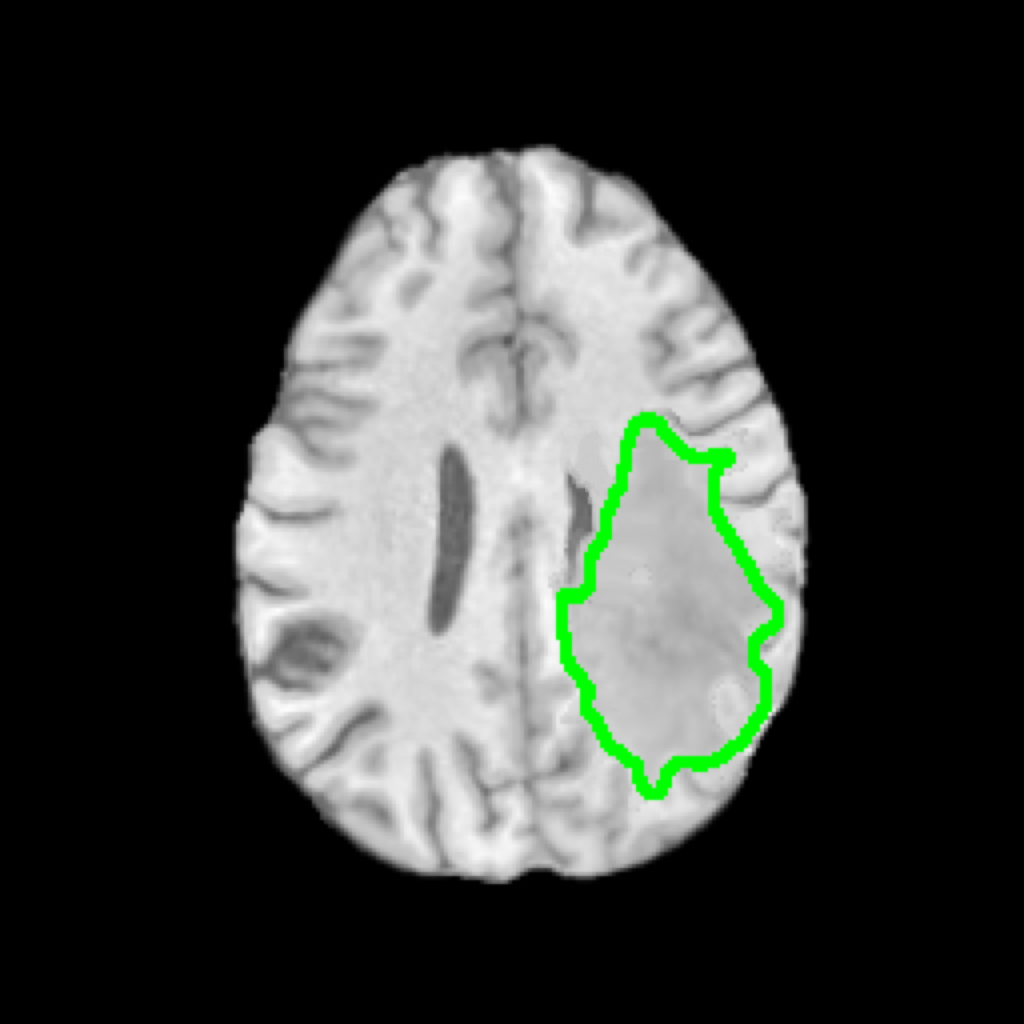}};

\draw[->,eccvblue!100,line width=0.8mm]
(\sx+5*\dx-0.2,\sy+0.8)--(\sx+5*\dx+0.13,\sy+0.2);


\node at (\sx-\dx,\metricA){DSC$\uparrow$/HD$\downarrow$};
\node at (\sx,\metricA){0.77/18.0};
\node at (\sx+\dx,\metricA){0.76/18.4};
\node at (\sx+2*\dx,\metricA){0.77/18.3};
\node at (\sx+3*\dx,\metricA){0.73/24.0};
\node at (\sx+4*\dx,\metricA){0.63/19.3};
\node at (\sx+5*\dx,\metricA){\textbf{0.85/16.4}};

\node at (\sx-\dx,\metricB){PSNR$\uparrow$/SSIM$\uparrow$};
\node at (\sx,\metricB){\textbf{29.94}/0.892};
\node at (\sx+\dx,\metricB){29.32/\textbf{0.894}};
\node at (\sx+2*\dx,\metricB){29.94/0.892};
\node at (\sx+3*\dx,\metricB){29.92/0.890};
\node at (\sx+4*\dx,\metricB){28.65/0.873};
\node at (\sx+5*\dx,\metricB){29.29/0.892};

\end{tikzpicture}
}

\caption{\textbf{An edge case of aggressive lesion progression.}
This example illustrates a case in which aggressive lesion progression causes \emph{structural distortion} of the originally normal ventricle.
Only \texttt{PDF} successfully captures this severe progression pattern.}

\label{fig:comp_aggre}

\end{figure}
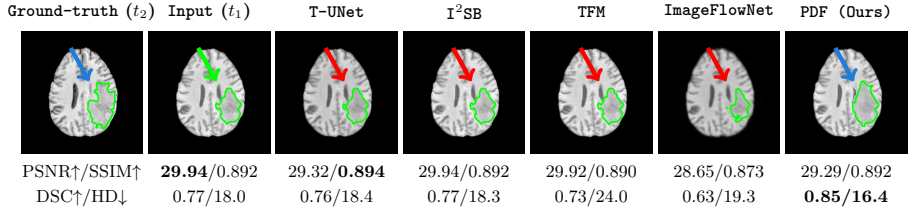

\paragraph{\textbf{Notes on evaluation metrics and performance interpretation.}}
We observe that image-level metrics (PSNR, SSIM, L1) may not fully reflect the lesion progression performance, as lesion changes are typically small compared to the whole brain region. Therefore, lesion-level metrics (DSC and HD) are more representative, as they focus on lesion regions in particular. 
As shown in Fig.~\ref{fig:comp_aggre}, although all models achieve satisfactory image-level performance, our method demonstrates much stronger progression modeling and more accurately captures aggressive tumor progression (DSC = \emph{0.85}), including the ventricle distortion, whereas other models struggle to model such pronounced structural changes.

\begin{table}[t] 
\centering
\normalsize
\setlength{\tabcolsep}{0.2mm}
\fontsize{9}{12}\selectfont
\resizebox{\linewidth}{!}{
\begin{tabular}{cc c cccc c cccc c cccc}
\toprule
\multicolumn{2}{c}{\textbf{Component}} &
& \multicolumn{4}{c}{\textbf{UCSF (Tumor)}} &
& \multicolumn{4}{c}{\textbf{LUMIERE (Tumor)}} & 
& \multicolumn{4}{c}{\textbf{LMSLS (MS)}} \\ [0.05cm]

 \cline{1-2}
\cline{4-7}
\cline{9-12}
\cline{14-17} \\ [-0.3cm]

\multicolumn{1}{c}{\fontsize{9}{12}\selectfont ~\textbf{Morph}~}
& \multicolumn{1}{c}{\fontsize{9}{12}\selectfont ~\textbf{Intensity}~} &~
& \multicolumn{1}{c}{\fontsize{9}{12}\selectfont \textbf{DSC} $\uparrow$}
& \multicolumn{1}{c}{\fontsize{9}{12}\selectfont \textbf{HD} $\downarrow$}
& \multicolumn{1}{c}{\fontsize{9}{12}\selectfont \textbf{PSNR} $\uparrow$}
& \multicolumn{1}{c}{\fontsize{9}{12}\selectfont \textbf{SSIM} $\uparrow$} &~
& \multicolumn{1}{c}{\fontsize{9}{12}\selectfont \textbf{DSC} $\uparrow$}
& \multicolumn{1}{c}{\fontsize{9}{12}\selectfont \textbf{HD} $\downarrow$}
& \multicolumn{1}{c}{\fontsize{9}{12}\selectfont \textbf{PSNR} $\uparrow$}
& \multicolumn{1}{c}{\fontsize{9}{12}\selectfont \textbf{SSIM} $\uparrow$} &~
& \multicolumn{1}{c}{\fontsize{9}{12}\selectfont \textbf{DSC} $\uparrow$}
& \multicolumn{1}{c}{\fontsize{9}{12}\selectfont \textbf{HD} $\downarrow$}
& \multicolumn{1}{c}{\fontsize{9}{12}\selectfont \textbf{PSNR} $\uparrow$}
& \multicolumn{1}{c}{\fontsize{9}{12}\selectfont \textbf{SSIM} $\uparrow$} \\
\midrule

\checkmark  &         &      & 0.750  & 26.038  & 31.343  & 0.920  & & 0.532  & 34.066  & \textbf{37.216}  & \textbf{0.958} & & 0.694  & 21.855  & \textbf{29.359}  & 0.900  \\
 &  \checkmark        &      & 0.749  & \textbf{25.806}  & \textbf{31.383}  & 0.920 & & 0.529  & 35.143  & 37.025  & 0.957 & & 0.690  & 21.749  & 29.307  & 0.900  \\
\checkmark    & \checkmark & & \textbf{0.752}  & 25.952  & 31.364  & \textbf{0.920} & & \textbf{0.534}  & \textbf{33.672}  & 37.075  & 0.957 & & \textbf{0.696}  & \textbf{20.867}  & 29.314  & \textbf{0.900} \\

\bottomrule
\end{tabular}
}  
\vspace{0.25cm}
\caption{\textbf{Ablation of disentangled morphology and intensity evolutions across three longitudinal datasets.} 
We evaluate morphology-only, intensity-only, and joint models to demonstrate the complementary strengths of the two components.}
\label{tab:ablation_disentangle}

\end{table}

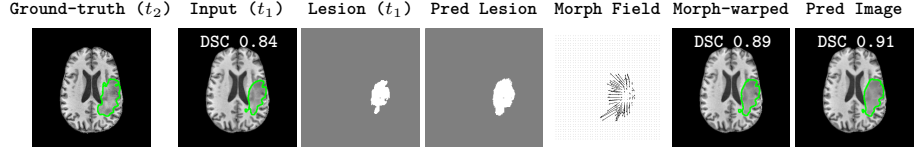
\begin{figure*}[t]
\centering
\fontsize{6.5}{10}\selectfont
\setlength{\tabcolsep}{1pt}
\renewcommand{\arraystretch}{1.1}
\resizebox{\linewidth}{!}{
\begin{tabular}{ccccccc}

{\ttfamily\bfseries Ground-truth ($t_2$)}
& {\ttfamily\bfseries Input ($t_1$)}
& {\ttfamily\bfseries Lesion ($t_1$)}
& {\ttfamily\bfseries Pred Lesion}
& {\ttfamily\bfseries Morph Field}
& {\ttfamily\bfseries Morph-warped}
& {\ttfamily\bfseries Pred Image} \\[2pt]

\begin{minipage}{0.12\linewidth}
  \centering
  \includegraphics[width=\linewidth]{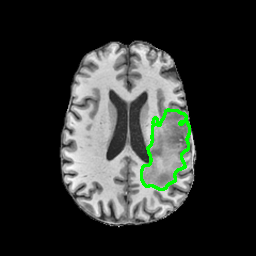}
\end{minipage}
&
\begin{minipage}{0.12\linewidth}
  \centering
  \setlength{\unitlength}{1\linewidth}
  \begin{picture}(1,1)
    \put(0,0){\includegraphics[width=\linewidth]{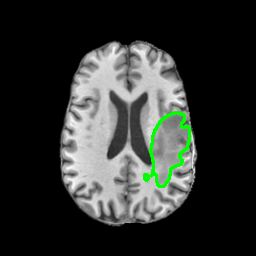}}
    \put(0.19,0.85){\color{white}\ttfamily\bfseries DSC 0.84}
  \end{picture}
\end{minipage}
&
\begin{minipage}{0.12\linewidth}
  \centering
  \includegraphics[width=\linewidth]{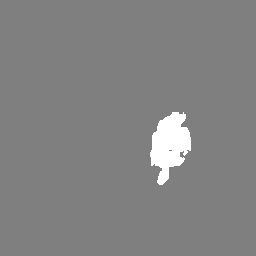}
\end{minipage}
&
\begin{minipage}{0.12\linewidth}
  \centering
  \includegraphics[width=\linewidth]{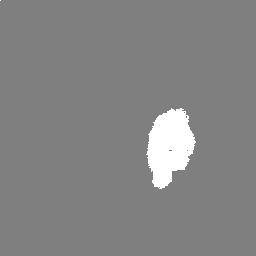}
\end{minipage}
&
\begin{minipage}{0.12\linewidth}
  \centering
  \includegraphics[width=\linewidth]{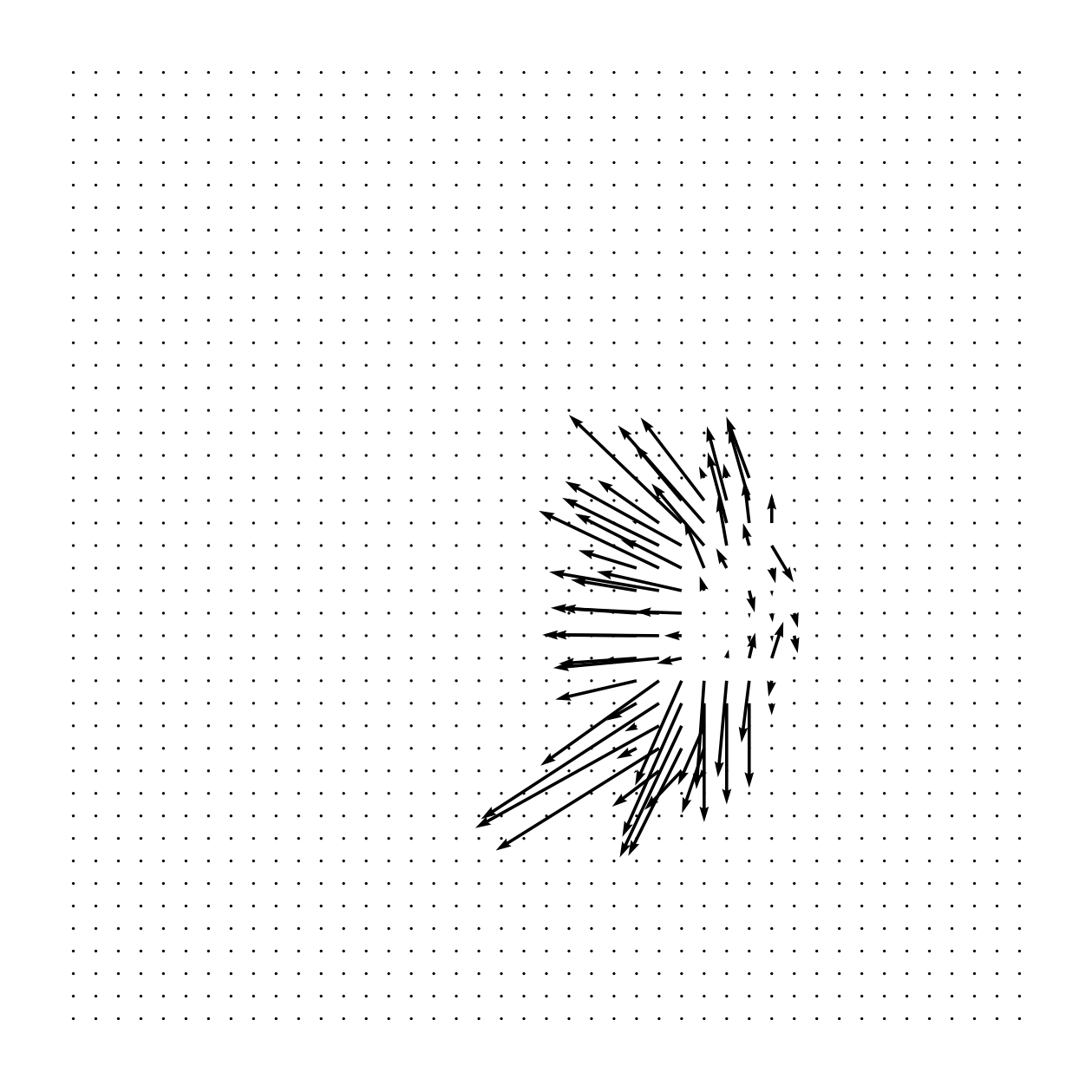}
\end{minipage}
&
\begin{minipage}{0.12\linewidth}
  \centering
  \setlength{\unitlength}{1\linewidth}
  \begin{picture}(1,1)
    \put(0,0){\includegraphics[width=\linewidth]{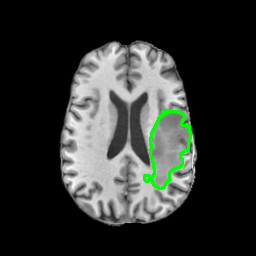}}
    \put(0.19,0.85){\color{white}\ttfamily\bfseries DSC 0.89}
  \end{picture}
\end{minipage}
&
\begin{minipage}{0.12\linewidth}
  \centering
  \setlength{\unitlength}{1\linewidth}
  \begin{picture}(1,1)
    \put(0,0){\includegraphics[width=\linewidth]{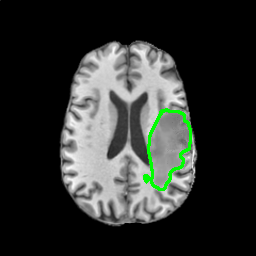}}
    \put(0.19,0.85){\color{white}\ttfamily\bfseries DSC 0.91}
  \end{picture}
\end{minipage}
\end{tabular}
}
\caption{\textbf{Qualitative visualization of disentangled prediction.}
From left to right: ground-truth and input images with lesions; original lesion at $t_1$ and predicted lesion mask, estimated morph field, morph-warped image, intensity-evolved final prediction.}
\label{fig:disentangle}
\end{figure*}

\subsection{Ablation Studies}

\paragraph{\textbf{Disentangled Components.}}
We perform ablation studies on the disentangled morphology and intensity evolution modeling by enabling each separately and comparing them with the jointly trained full model. Since the intensity branch requires a predicted lesion mask, we use the mask from the morphology branch while disabling structural deformation. 

Tab.~\ref{tab:ablation_disentangle} reports the quantitative results. 
The morphology branch alone improves lesion-level metrics, particularly DSC, highlighting the importance of structural deformation modeling. 
The intensity branch alone yields competitive PSNR and SSIM, reflecting strong appearance modeling. 
When combined, the full model achieves the best or near-best performance across datasets. 
In particular, on UCSF and LMSLS, the joint model improves DSC and reduces HD compared to either single component, indicating that morphology and intensity modeling provide complementary benefits.

Fig.~\ref{fig:disentangle} provides qualitative visualizations of the disentangled components. 
Applying the morphology branch first produces a structurally aligned prediction that yields structurally consistent evolution, improving DSC from 0.84 to 0.89. 
The subsequent intensity evolution refines lesion boundaries and internal textures, further increasing DSC to 0.91. The structural alignment corrects lesion-induced spatial shifts, while intensity modeling captures residual appearance changes, together producing more accurate longitudinal lesion trajectories.

\begin{table}[t]
\centering
\normalsize
\setlength{\tabcolsep}{0.2mm}
\fontsize{9}{13}\selectfont
\resizebox{\linewidth}{!}{
\begin{tabular}{l cccc c cccc c cccc}
\toprule
\multirow{2}{*}{\textbf{Loss}} 
& \multicolumn{4}{c}{\textbf{UCSF (Tumor)}} &
& \multicolumn{4}{c}{\textbf{LUMIERE (Tumor)}} &
& \multicolumn{4}{c}{\textbf{LMSLS (MS)}} \\ [0.05cm]

\cline{2-5}
\cline{7-10}
\cline{12-15} \\ [-0.3cm]

& \multicolumn{1}{c}{\fontsize{9}{13}\selectfont \textbf{DSC} $\uparrow$}
& \multicolumn{1}{c}{\fontsize{9}{13}\selectfont \textbf{HD} $\downarrow$}
& \multicolumn{1}{c}{\fontsize{9}{13}\selectfont \textbf{PSNR} $\uparrow$}
& \multicolumn{1}{c}{\fontsize{9}{13}\selectfont \textbf{SSIM} $\uparrow$} &~
& \multicolumn{1}{c}{\fontsize{9}{13}\selectfont \textbf{DSC} $\uparrow$}
& \multicolumn{1}{c}{\fontsize{9}{13}\selectfont \textbf{HD} $\downarrow$}
& \multicolumn{1}{c}{\fontsize{9}{13}\selectfont \textbf{PSNR} $\uparrow$}
& \multicolumn{1}{c}{\fontsize{9}{13}\selectfont \textbf{SSIM} $\uparrow$} &~
& \multicolumn{1}{c}{\fontsize{9}{13}\selectfont \textbf{DSC} $\uparrow$}
& \multicolumn{1}{c}{\fontsize{9}{13}\selectfont \textbf{HD} $\downarrow$}
& \multicolumn{1}{c}{\fontsize{9}{13}\selectfont \textbf{PSNR} $\uparrow$}
& \multicolumn{1}{c}{\fontsize{9}{13}\selectfont \textbf{SSIM} $\uparrow$} \\
\midrule

w/o $\mathcal{L}_{\text{PDE}}~~$  & 0.745   & 26.216    & 31.359   & 0.920  & & 0.524   & 34.211   & 37.137   & 0.953 & & 0.690   & 22.337   & 29.312  & 0.898 \\
w/ $\mathcal{L}_{\text{PDE}}$    & \textbf{0.752}   & \textbf{25.952}    & \textbf{31.364}   & \textbf{0.920}  & & \textbf{0.534}   & \textbf{33.672}   & \textbf{37.216}   & \textbf{0.957} & & \textbf{0.696}   & \textbf{20.867}   & \textbf{29.314}  & \textbf{0.900} \\

\bottomrule
\end{tabular}
}
\vspace{0.25cm}
\caption{\textbf{Ablation of PDE-regularized loss across three longitudinal datasets.} 
We compare models trained with and without $\mathcal{L}_{\text{PDE}}$. 
The model with $\mathcal{L}_{\text{PDE}}$ consistently improves lesion-level metrics while maintaining image-level performance.}
\label{tab:ablation_physics}

\end{table}


\begin{figure*}[t]
\centering
\fontsize{6.5}{10}\selectfont
\setlength{\tabcolsep}{1pt}
\renewcommand{\arraystretch}{1.1}
\resizebox{\linewidth}{!}{
\begin{tabular}{ccccccc}

{\ttfamily\bfseries Ground-truth}
& {\ttfamily\bfseries Input}
& 
& {\ttfamily\bfseries Velocity}
& {\ttfamily\bfseries Projection}
& {\ttfamily\bfseries Pred Image}
& {\ttfamily\bfseries Lesion Region} \\[2pt]


\begin{minipage}{0.12\linewidth}
\centering
\includegraphics[width=\linewidth]{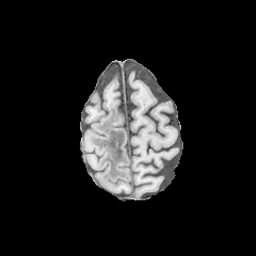}
\end{minipage}
&
\begin{minipage}{0.12\linewidth}
\centering
\includegraphics[width=\linewidth]{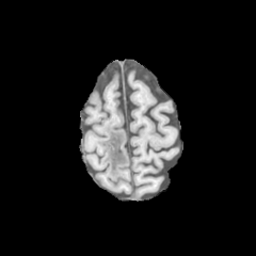}
\end{minipage}
&
{\fontsize{6.5}{10}\selectfont \texttt{~w/o $\mathcal{L}_{\text{PDE}}$~}}
&
\begin{minipage}{0.12\linewidth}
\centering
\includegraphics[width=\linewidth]{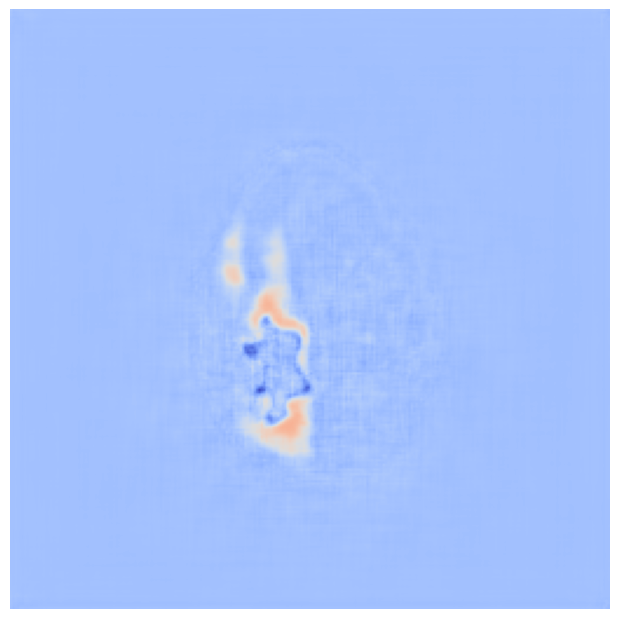}
\end{minipage}
&
\begin{minipage}{0.12\linewidth}
\centering
\includegraphics[width=\linewidth]{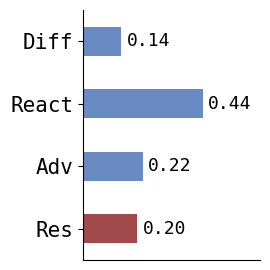}
\end{minipage}
&
\begin{minipage}{0.12\linewidth}
\centering
\includegraphics[width=\linewidth]{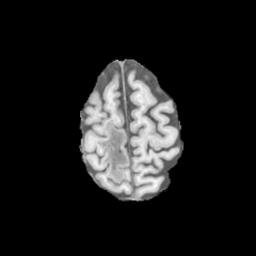}
\end{minipage}
&
\begin{minipage}{0.12\linewidth}
\centering
\setlength{\unitlength}{1\linewidth}
\begin{picture}(1,1)
\put(0,0){\includegraphics[width=\linewidth]{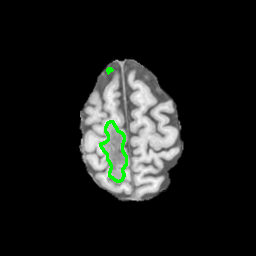}}
\put(0.15,0.85){\color{white}\ttfamily\bfseries DSC 0.69}
\end{picture}
\end{minipage}
\\[20pt]


\begin{minipage}{0.12\linewidth}
\centering
\includegraphics[width=\linewidth]{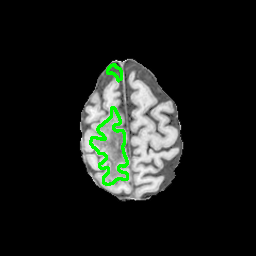}
\end{minipage}
&
\begin{minipage}{0.12\linewidth}
\centering
\setlength{\unitlength}{1\linewidth}
\begin{picture}(1,1)
\put(0,0){\includegraphics[width=\linewidth]{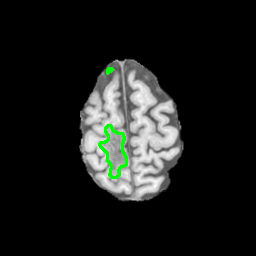}}
\put(0.15,0.85){\color{white}\ttfamily\bfseries DSC 0.64}
\end{picture}
\end{minipage}
&
{\fontsize{6.5}{10}\selectfont \texttt{w/ $\mathcal{L}_{\text{PDE}}$}}
&
\begin{minipage}{0.12\linewidth}
\centering
\includegraphics[width=\linewidth]{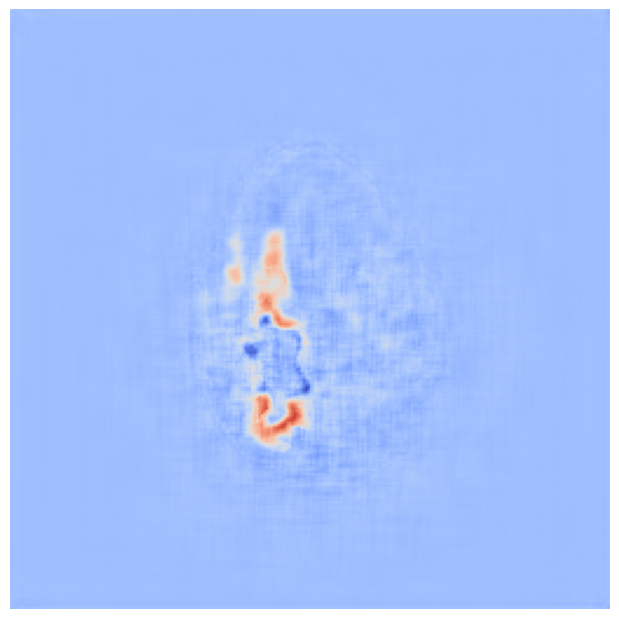}
\end{minipage}
&
\begin{minipage}{0.12\linewidth}
\centering
\includegraphics[width=\linewidth]{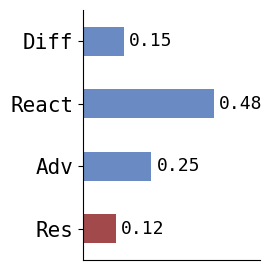}
\end{minipage}
&
\begin{minipage}{0.12\linewidth}
\centering
\includegraphics[width=\linewidth]{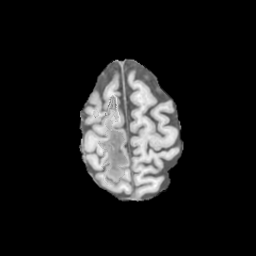}
\end{minipage}
&
\begin{minipage}{0.12\linewidth}
\centering
\setlength{\unitlength}{1\linewidth}
\begin{picture}(1,1)
\put(0,0){\includegraphics[width=\linewidth]{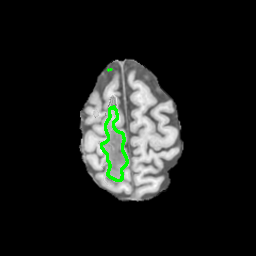}}
\put(0.15,0.85){\color{white}\ttfamily\bfseries DSC 0.73}
\end{picture}
\end{minipage}

\end{tabular}
}
\caption{\textbf{Effect of PDE-regularized loss on longitudinal prediction.}
Left to right: ground-truth and input images; predicted morphology evolution velocity; energy projection within physics span and its residual; predicted images and lesion regions with and without $\mathcal{L}_{\text{PDE}}$.
$\mathcal{L}_{\text{PDE}}$ reduces residual energy and improves lesion evolution.}

\label{fig:justification}

\end{figure*}
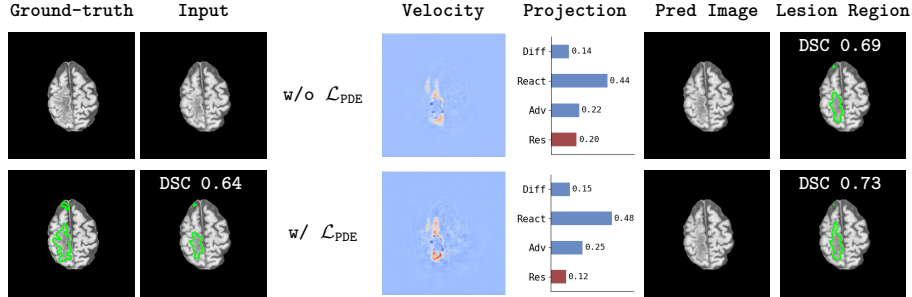

\paragraph{\textbf{PDE-Regularized Loss.}}
We also evaluate the effect of the proposed PDE-regularized loss by comparing models trained with and without $\mathcal{L}_{\text{PDE}}$ (\cref{eq:pde_loss}).

Tab.~\ref{tab:ablation_physics} summarizes the quantitative results. 
Across all three datasets, incorporating $\mathcal{L}_{\text{PDE}}$ consistently improves lesion-level metrics. 
On UCSF, DSC increases from 0.745 to 0.752 while HD decreases from 26.216 to 25.952. 
On LUMIERE, both DSC and HD show improvements, and on LMSLS the gain is more pronounced, with DSC increasing from 0.690 to 0.696 and HD decreasing from 22.337 to 20.867. 
These results suggest that PDE regularization improves structural consistency in longitudinal prediction.

Fig.~\ref{fig:justification} further illustrates the effect of $\mathcal{L}_{\text{PDE}}$ on a representative case. 
The first two columns show the ground-truth and input images, with their lesion regions outlined. 
The third column visualizes the predicted mask-evolution velocity with and without PDE regularization. 
The fourth column presents the energy decomposition of the velocity projected onto the predefined PDE basis and the corresponding residual ratio. 
Without $\mathcal{L}_{\text{PDE}}$, the residual energy is 0.20, indicating that a substantial portion of the predicted dynamics lies outside the physics-based span. 
With PDE regularization, the residual decreases to 0.12, showing improved alignment with the underlying physical basis. 
Consistently, the predicted lesion DSC increases from 0.69 to 0.73, with better shape agreement with the ground truth. 
These results suggest that $\mathcal{L}_{\text{PDE}}$ promotes physically consistent evolution and improves structural accuracy.

\paragraph{\textbf{Key Parameter.}}

\begin{table}[t]
\centering
\normalsize
\setlength{\tabcolsep}{0.2mm}
\fontsize{9}{13}\selectfont
\resizebox{\linewidth}{!}{
\begin{tabular}{l cccc c cccc c cccc}
\toprule
\multirow{2}{*}{\textbf{Weight}~~} 
& \multicolumn{4}{c}{\textbf{UCSF}} & 
& \multicolumn{4}{c}{\textbf{LUMIERE}} &  
& \multicolumn{4}{c}{\textbf{LMSLS}} \\ [0.05cm]

\cline{2-5} 
\cline{7-10} 
\cline{12-15}  \\ [-0.3cm]

& \multicolumn{1}{c}{\fontsize{9}{13}\selectfont \textbf{DSC} $\uparrow$}
& \multicolumn{1}{c}{\fontsize{9}{13}\selectfont \textbf{HD} $\downarrow$}
& \multicolumn{1}{c}{\fontsize{9}{13}\selectfont \textbf{PSNR} $\uparrow$}
& \multicolumn{1}{c}{\fontsize{9}{13}\selectfont \textbf{SSIM} $\uparrow$} &~
& \multicolumn{1}{c}{\fontsize{9}{13}\selectfont \textbf{DSC} $\uparrow$}
& \multicolumn{1}{c}{\fontsize{9}{13}\selectfont \textbf{HD} $\downarrow$}
& \multicolumn{1}{c}{\fontsize{9}{13}\selectfont \textbf{PSNR} $\uparrow$}
& \multicolumn{1}{c}{\fontsize{9}{13}\selectfont \textbf{SSIM} $\uparrow$} &~
& \multicolumn{1}{c}{\fontsize{9}{13}\selectfont \textbf{DSC} $\uparrow$}
& \multicolumn{1}{c}{\fontsize{9}{13}\selectfont \textbf{HD} $\downarrow$}
& \multicolumn{1}{c}{\fontsize{9}{13}\selectfont \textbf{PSNR} $\uparrow$}
& \multicolumn{1}{c}{\fontsize{9}{13}\selectfont \textbf{SSIM} $\uparrow$} \\
\midrule

$\lambda = 1$       & 0.745 & 26.224 & \textbf{31.482} & 0.920 &
                    & 0.526 & 35.241 & 37.134 & 0.925 &
                    & 0.691 & 21.634 & 29.307 & 0.898 \\
                    
$\lambda = 0.1$     & 0.746 & \textbf{25.635} & 31.321 & 0.920 &
                    & 0.529 & 34.128 & 37.163 & 0.936 &
                    & 0.693 & 21.402 & \textbf{29.322} & 0.898 \\

$\lambda = 0.01$    & \textbf{0.752} & 25.952 & 31.364 & \textbf{0.920} &
                    & \textbf{0.534} & \textbf{33.672} & \textbf{37.216} & \textbf{0.957}  &
                    & \textbf{0.696} & \textbf{20.867} & 29.314 & \textbf{0.900} \\
\bottomrule
\end{tabular} 
}
\vspace{0.25cm}
\caption{\textbf{Ablation of key parameter across three longitudinal brain MRI datasets.}
We evaluate different values of $\lambda$ across longitudinal brain MRI datasets. 
}
\label{tab:ablation_param}

\end{table}

We further investigate the influence of the PDE regularization weight $\lambda$ in the morphological loss (Eq.~\eqref{eq:morph_loss}), where $\lambda$ controls the balance between data-driven flow matching and physics-grounded PDE regularization. 

As shown in Tab.~\ref{tab:ablation_param}, moderate regularization ($\lambda=0.01$) consistently achieves the best overall lesion-level performance across datasets. 
Larger values ($\lambda=0.1$ and $\lambda=1$) slightly degrade DSC and HD, suggesting that overly strong physical constraints may hinder accurate deformation modeling. 
These results indicate that appropriately balancing data-driven learning and physics-based regularization is crucial for stable and effective longitudinal forecasting.

\section{Conclusion}
We presented \textbf{PDF}, a physics-grounded disentangled flow matching framework for longitudinal brain disease progression. 
By disentangling disease morphological evolution from image intensity evolution, our approach enables structured and interpretable modeling of disease progression. 
Incorporating a PDE-regularized loss grounded in lesion growth dynamics further constrains the predicted trajectories within a biologically meaningful diffusion-reaction-advection formulation, improving physical plausibility and longitudinal consistency. 
Experiments across three longitudinal datasets demonstrate that \textbf{PDF} achieves state-of-the-art performance while producing more coherent lesion evolution patterns. 
These results highlight the importance of integrating disentangled modeling and physics-grounded constraints for reliable longitudinal disease forecasting. 


\paragraph{\textbf{Acknowledgements.}} We gratefully acknowledge the support of NVIDIA Corporation for providing the  NVIDIA RTX PRO 6000 GPU used in this research.

%
%
\bibliographystyle{splncs04}
\bibliography{main}

\clearpage

\begin{center}
{\Large \hspace*{-1.3cm}
    \titleicon
    \hspace{0.06cm} 
    \textbf{Physics-Grounded Disentangled Flow Modeling\\\vspace{0.15cm}for Brain Disease Progression Trajectory}} \\ \vspace{0.4cm}
{\large \textit{Supplementary Material}}
\end{center}

\appendix
\renewcommand{\theHsection}{A\arabic{section}}
\setcounter{figure}{0}
\setcounter{equation}{0}
\setcounter{section}{0}

\renewcommand\thefigure{\thesection.\arabic{figure}}
\renewcommand\thefigure{\thesection.\arabic{table}}
\renewcommand\theequation{\thesection.\arabic{equation}}
\counterwithin{figure}{section}
\counterwithin{table}{section}
\counterwithin{equation}{section}

\vspace{0.4cm}
\noindent \textbf{This supplementary material includes:}

\begin{table}
\vspace{-0.5cm}
\begin{tabular}{ cll } 
{\bf\ref{sec:data_preprocess}} & - & \textbf{Dataset Preprocessing} \\ 
 {\bf\ref{sec:dataset_projection}} & - & \textbf{Dataset-Level PDE-Based Projection} \\ 
 {\bf\ref{sec:eval_metrics}} & - & \textbf{Evaluation Metrics} \\ 
{\bf\ref{sec:impl_details}} & - & \textbf{Further Model Details} \\ 
{\bf\ref{sec:future}} & - & \textbf{Limitations and Future Directions}
\end{tabular}
\vspace*{-0.8cm}
\end{table}


\section{Dataset Preprocessing}
\label{sec:data_preprocess}
Longitudinal image datasets often suffer from spatial misalignment among images acquired at different time points. This phenomenon is nearly inevitable in clinical imaging, as even minor changes in patient positioning or acquisition angle can disrupt exact pixel-wise alignment across visits. To address this issue, we spatially align images within each longitudinal series during preprocessing stage.

Specifically, we apply affine registration to align scans across time points using the ANTs registration toolkit \cite{scheufele2020image}. In addition to spatial alignment, we further normalize image intensity distributions using histogram matching, which mitigates variations caused by scanner differences or acquisition settings. For each longitudinal series, we align the intensity distribution of later scans to the first time point, ensuring consistent intensity statistics across the sequence.

\subsection{UCSF Dataset}

The UCSF~\cite{fields2024university} dataset provides longitudinal brain scans that have already been spatially registered across time points. Therefore, no additional spatial registration is required. However, intensity variations still exist across scans within each longitudinal series. To improve temporal consistency, we apply histogram matching to align the intensity distribution of each follow-up scan to that of the baseline scan.

\subsection{LUMIERE Dataset}

For the LUMIERE~\cite{suter2022lumiere} dataset, we use scans from the ``DeepBraTumIA'' folders, which were registered to a common atlas. However, this atlas-based registration does not sufficiently align scans within each longitudinal series. To address this issue, we perform additional registration using the ANTs Python toolkit \cite{scheufele2020image}. Specifically, we apply affine registration followed by diffeomorphic registration with a multi-resolution schedule of [4, 2, 1] iterations to align each scan to the first scan in the series.
After spatial alignment, we further normalize intensity distributions within each longitudinal series by performing histogram matching. For each patient, all follow-up scans are matched to the histogram of the baseline scan.

\subsection{LMSLS Dataset}

The LMSLS~\cite{carass2017longitudinal} dataset provides longitudinal scans that have already been spatially registered. Therefore, we do not perform additional spatial registration. To reduce intensity variation across time points, we apply histogram matching within each longitudinal series, aligning the intensity distribution of each follow-up scan to the baseline scan.

\begin{figure}[t]
\centering
\includegraphics[width=0.95\linewidth]{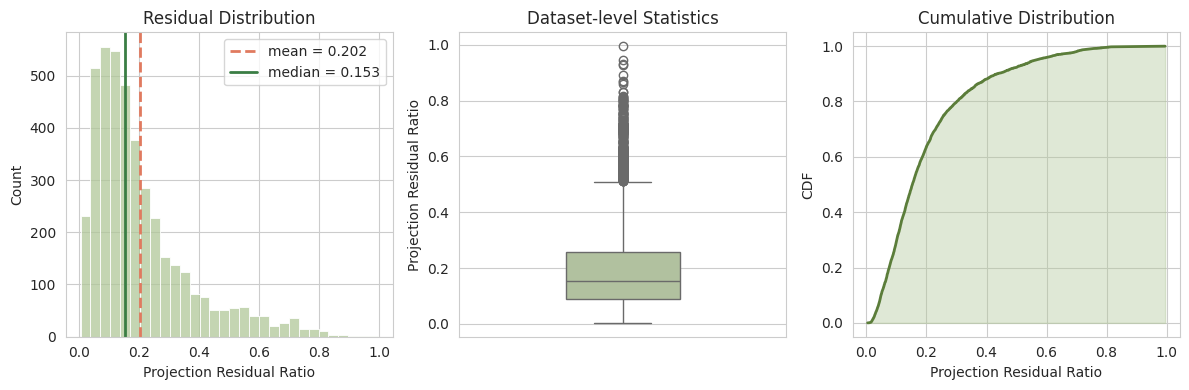}
\caption{\textbf{Dataset-level statistics of projection residuals.}
The projection residual ratio $\epsilon$ quantifies the discrepancy between the real lesion evolution velocity and its projection onto the diffusion-reaction-advection physics span.
\textbf{Left:} histogram of residual ratios across the UCSF dataset with mean and median indicated.
\textbf{Middle:} boxplot summarizing the dataset-level statistics.
\textbf{Right:} cumulative distribution function (CDF) showing that most samples have small projection residuals.}
\label{fig:statistics}
\end{figure}

\section{Dataset-Level PDE-Based Projection}
\label{sec:dataset_projection}

To validate the physical plausibility of the PDE-based modeling introduced in Sec.~\ref{subsubsec:pde_progression}, we perform a dataset-level analysis of the projection error onto the diffusion-reaction-advection physics span.

We conduct this analysis on the UCSF dataset, which provides expert-annotated lesion masks that allow reliable estimation of lesion progression dynamics. For each longitudinal pair of scans $(t_1, t_2)$, we follow the procedure described in Sec.~\ref{subsubsec:pde_progression} to estimate the PDE components and compute the projection residual that measures the discrepancy between the observed lesion evolution and its projection onto the physics span.

The dataset-level statistics are summarized in Fig.~\ref{fig:statistics}. The histogram shows that projection residuals are concentrated near zero, with a mean of 0.202 and a median of 0.153, indicating that most longitudinal pairs can be well approximated by the diffusion-reaction-advection dynamics.

The boxplot further confirms that the majority of samples fall within a relatively small residual range, with most values below 0.3. The cumulative distribution function (CDF) provides a global view of this trend, showing that a large proportion of samples exhibit small projection residuals.

These observations suggest that real lesion evolution largely lies within the diffusion-reaction-advection physics span, providing empirical support for the validity of the proposed PDE-based formulation.

\section{Evaluation Metrics}
\label{sec:eval_metrics}

We evaluate the performance of longitudinal prediction from both image similarity and lesion-region consistency perspectives.

\subsection{Image Similarity}
We measure the similarity between the real future image $x_{t_2}$ and the predicted image $\hat{x}_{t_2}$ using peak signal-to-noise ratio (PSNR) and structural similarity index (SSIM), two widely used metrics for image-to-image tasks.

\noindent - PSNR is a normalized form of the mean squared error (MSE) that accounts for the dynamic range of image intensities. It is defined as
\begin{equation}
\text{PSNR}(x_a, x_b) = 10 \lg \left( \frac{R^2}{\text{MSE}(x_a, x_b)} \right),
\end{equation}
where $R$ denotes the dynamic range of the image intensity values and
\begin{equation}
\text{MSE}(x_a, x_b) = \frac{1}{HW}\sum_{h=1}^{H}\sum_{w=1}^{W}
\left(x_a^{(h,w)} - x_b^{(h,w)}\right)^2 .
\end{equation}

\noindent - SSIM measures structural similarity by comparing luminance, contrast, and structural information between two images:
\begin{equation}
\text{SSIM}(x_a, x_b) =
\frac{(2\mu_a \mu_b + c_1)(2\sigma_{ab} + c_2)}
{(\mu_a^2 + \mu_b^2 + c_1)(\sigma_a^2 + \sigma_b^2 + c_2)},
\end{equation}
where $\mu_a$ and $\mu_b$ denote the mean intensities, $\sigma_a^2$ and $\sigma_b^2$ denote the variances, and $\sigma_{ab}$ denotes the covariance between the two images. The constants are defined as $c_1=(0.01R)^2$ and $c_2=(0.03R)^2$.

\subsection{Lesion Region Consistency}
To evaluate how accurately the predicted images capture lesion progression, we compute the Dice similarity coefficient (DSC) and Hausdorff distance (HD) between the predicted and ground-truth lesion masks.

\noindent - The Dice similarity coefficient between two binary masks $X$ and $Y$:
\begin{equation}
\text{DSC}(X,Y) = \frac{2|X \cap Y|}{|X| + |Y|}.
\end{equation}

\noindent - The Hausdorff distance measures the maximum boundary discrepancy between two regions:
\begin{equation}
\text{HD}(X,Y) =
\max \left\{
\sup_{x \in X} d(x,Y), \,
\sup_{y \in Y} d(X,y)
\right\},
\end{equation}
where $d(\cdot,\cdot)$ denotes the Euclidean distance between two sets.

These metrics jointly evaluate both the image-level reconstruction quality and the spatial accuracy of lesion evolution.

\section{Further Model Details}
\label{sec:impl_details}


To better preserve lesion-specific texture during intensity evolution, we incorporate high-frequency components extracted from the baseline image $I_{t_1}$ as additional guidance during refinement.

Specifically, we first isolate the lesion region using the lesion mask $M_{t_1}$. The lesion intensity image is obtained by
\begin{equation}
I^{\text{lesion}}_{t_1} = I_{t_1} \odot M_{t_1},
\end{equation}
where $\odot$ denotes element-wise multiplication.

We then extract high-frequency components from the lesion region using a Difference-of-Gaussians (DoG) filter. Let $G_{\sigma}$ denote a Gaussian smoothing operator with standard deviation $\sigma$. The high-frequency component is computed as
\begin{equation}
H_{t_1} = G_{\sigma_1}(I^{\text{lesion}}_{t_1}) - G_{\sigma_2}(I^{\text{lesion}}_{t_1}),
\end{equation}
where $\sigma_1 < \sigma_2$. This operation removes low-frequency intensity variations while retaining fine structural details and lesion textures.

During inference, the intensity flow matching module predicts a refined image $\hat{I}_{t_2}^{\text{intensity}}$. The extracted high-frequency component from the baseline scan is then injected to preserve texture details:
\begin{equation}
\hat{I}_{t_2} = \hat{I}_{t_2}^{\text{intensity}} + \lambda H_{t_1},
\end{equation}
where $\lambda$ controls the contribution of the high-frequency guidance.

This strategy helps maintain lesion-specific texture patterns that may otherwise be smoothed out during intensity evolution, improving visual fidelity and structural consistency of the predicted images.

\section{Limitations and Future Directions}
\label{sec:future}

\subsection{Limitations} 
Although \textbf{PDF} demonstrates strong performance and improved physical consistency in longitudinal lesion prediction, several limitations remain. \textit{First}, our model relies on the availability of lesion masks at the baseline time point. Inaccurate or noisy segmentations may propagate errors into the morphology evolution modeling and subsequently affect the predicted future images. While high-quality annotations are available in the datasets used in this work, improving robustness to imperfect lesion segmentation remains an important direction for future research.
\textit{Second}, \textbf{PDF} explicitly disentangles morphology evolution and intensity evolution using two separate flow matching models. Although this design improves interpretability and modeling flexibility, it introduces additional computational overhead compared with single-model approaches.
\textit{Furthermore}, in the current formulation, the PDE dynamics serve primarily as a regularization constraint that guides the learned morphology evolution, rather than being directly embedded into the generative process.

\subsection{Future Work}
Our future work will explore several directions to address these limitations. These include developing more robust lesion-aware representations that reduce dependence on precise, pixel-level segmentation, designing more efficient unified models that jointly capture morphology and intensity evolution, and more tightly integrating physics-based operators into the learning framework. Moreover, the proposed physics-guided formulation has the potential to generate physically consistent synthetic longitudinal trajectories, helping alleviate the scarcity of longitudinal medical imaging data and enabling more effective training in data-limited settings. We will further investigate and quantify the benefits of such generated data.

\end{document}